\documentclass[letterpaper, 10 pt, conference]{ieeeconf}  

\IEEEoverridecommandlockouts                              

\usepackage{graphicx}
\usepackage{amssymb}
\usepackage{amsmath}
\usepackage{multirow}
\usepackage{cleveref}
\usepackage{booktabs}
\usepackage{pifont}
\usepackage{makecell}
\usepackage{algorithm}
\usepackage{algpseudocode}
\let\labelindent\relax  
\usepackage{enumitem}
\usepackage{pifont}
\usepackage{url}

\usepackage{subcaption}
\usepackage[margin=1in]{geometry}

\title{\LARGE \bf
Language-Guided Open-World Anomaly Segmentation
}

\author{Klara Reichard$^{1,2}$, Nikolas Brasch$^{1}$, Nassir Navab$^{1}$,  Federico Tombari$^{1,5}$
\thanks{$^{1}$ Technical University of Munich, Germany.}%
\thanks{$^{2}$ BMW Group, Munich, Germany.}
}

\begin{document}

\maketitle
\thispagestyle{empty}
\pagestyle{empty}

\begin{abstract}
Open-world and anomaly segmentation methods seek to enable autonomous driving systems to detect and segment both known and unknown objects in real-world scenes. However, existing methods do not assign semantically meaningful labels to unknown regions, and distinguishing and learning representations for unknown classes remains difficult. While open-vocabulary segmentation methods show promise in generalizing to novel classes, they require a fixed inference vocabulary and thus cannot be directly applied to anomaly segmentation where unknown classes are unconstrained. We propose Clipomaly, the first CLIP-based open-world and anomaly segmentation method for autonomous driving. Our zero-shot approach requires no anomaly-specific training data and leverages CLIP’s shared image-text embedding space to both segment unknown objects and assign human-interpretable names to them. Unlike open-vocabulary methods, our model dynamically extends its vocabulary at inference time without retraining, enabling robust detection and naming of anomalies beyond common class definitions such as those in Cityscapes. Clipomaly achieves state-of-the-art performance on established anomaly segmentation benchmarks while providing interpretability and flexibility essential for practical deployment.
\end{abstract}
\begin{figure}[H]
    \centering

    \begin{subfigure}[b]{0.23\textwidth}
        \centering
        \includegraphics[width=\textwidth]{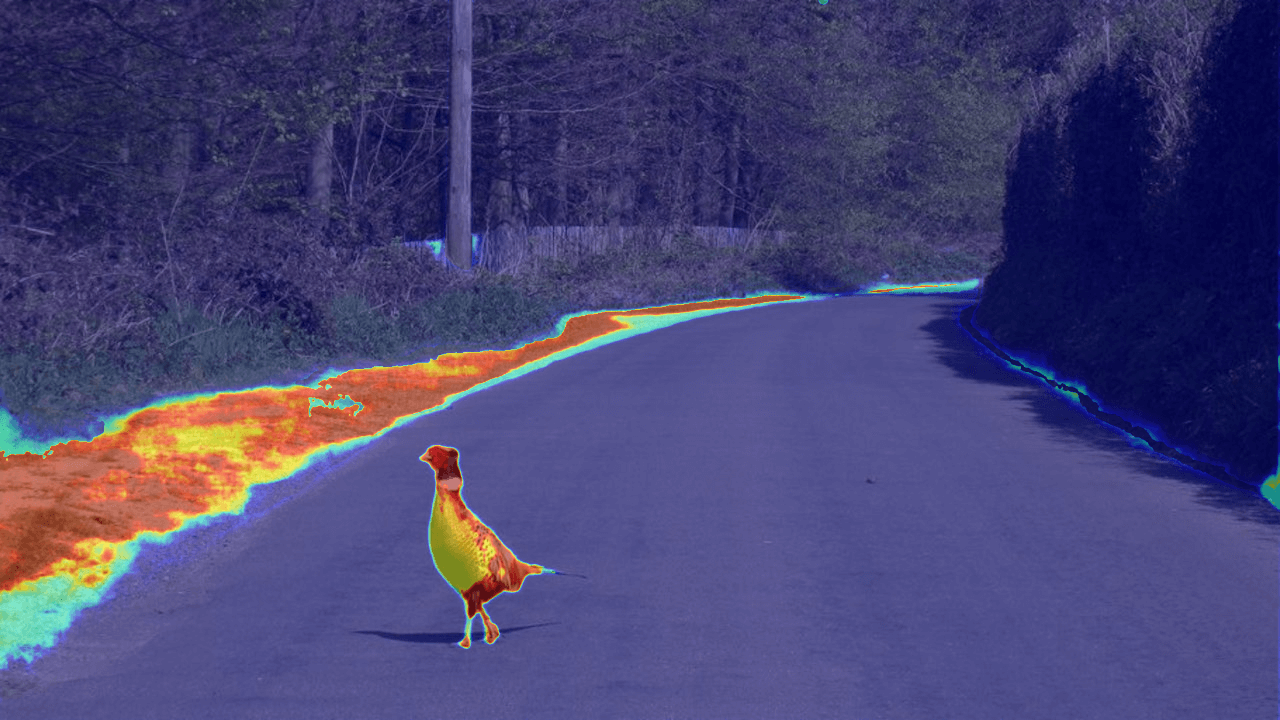}
        \caption{SOTA Unknowns}
    \end{subfigure}
    \hfill
    \begin{subfigure}[b]{0.23\textwidth}
        \centering
        \includegraphics[width=\textwidth]{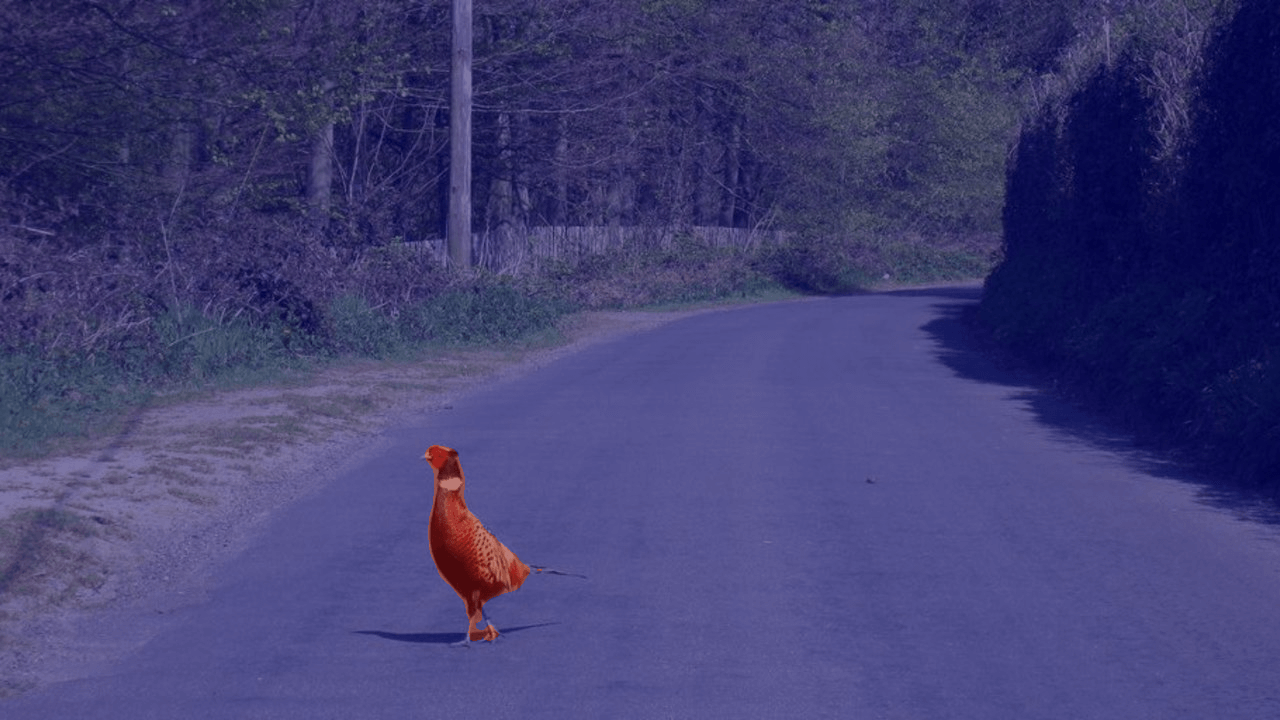}
        \caption{Ours Unknowns}
    \end{subfigure}
    \hfill
    \begin{subfigure}[b]{0.23\textwidth}
        \centering
        \includegraphics[width=\textwidth]{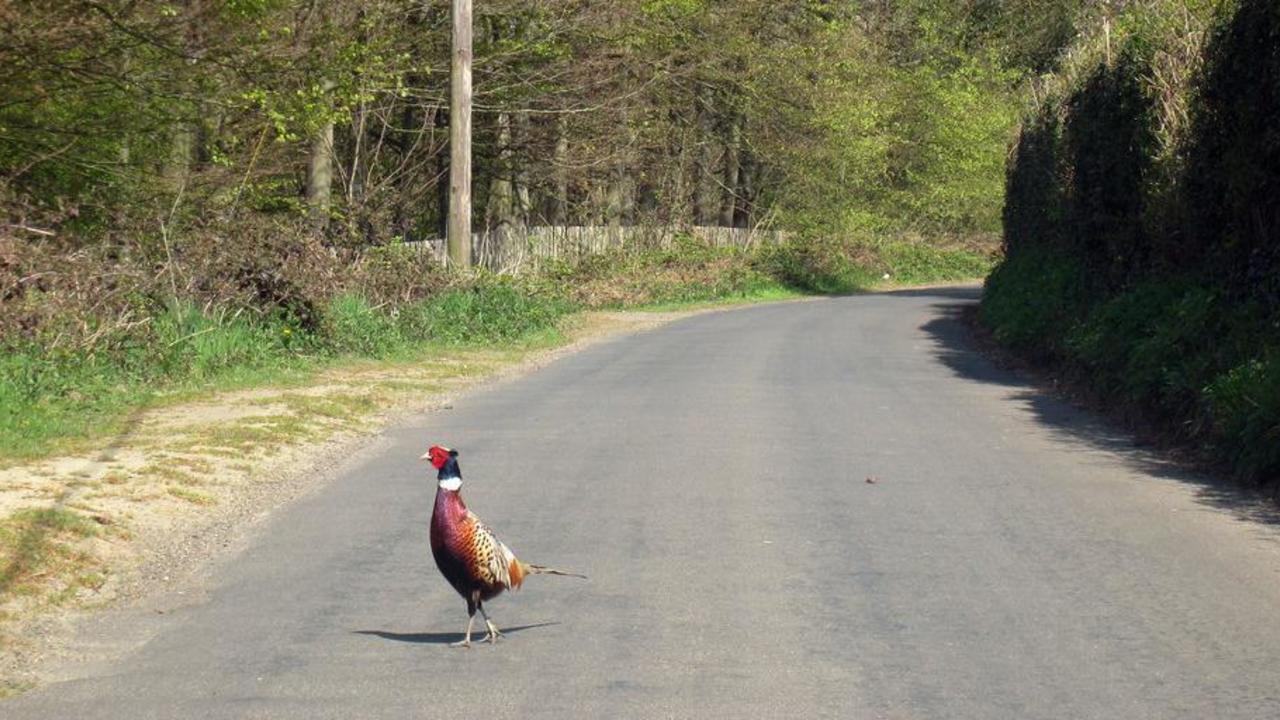}
        \caption{Input Image}
    \end{subfigure}
    \hfill
    \begin{subfigure}[b]{0.23\textwidth}
        \centering
        \includegraphics[width=\textwidth]{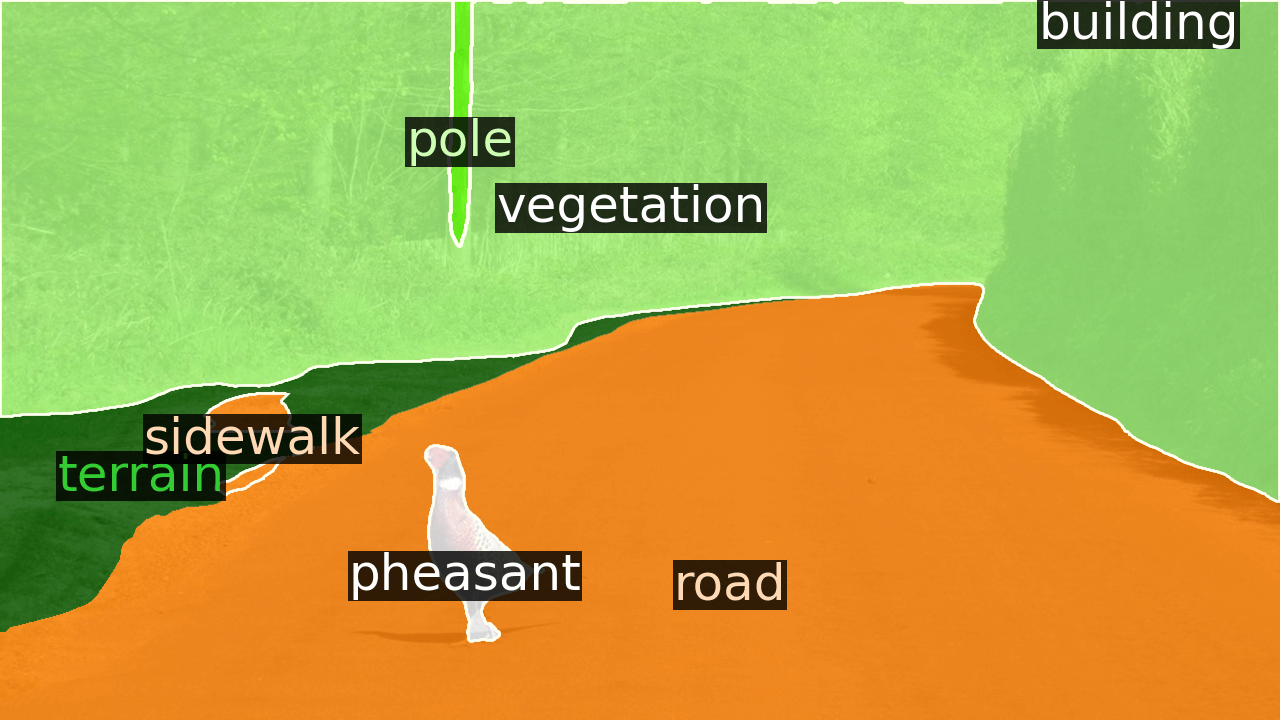}
        \caption{Ours (Open World)}
    \end{subfigure}

    \caption{Teaser: Our method produces accurate known-class and detailed unknown-class segmentations with meaningful semantic labels.} 
    \label{fig:teaser}
\end{figure}


\section{INTRODUCTION}
\label{sec:intro}

Scene understanding is a critical component of perception systems for autonomous driving. Traditionally, semantic segmentation in this domain has relied on closed-world assumptions, i.e., models are trained on a fixed set of known classes (e.g., from Cityscapes) and are expected to predict only from this predefined set. Many conventional methods, such as \cite{fullyconvolutionalnetworkssemantic}, operate under this closed-set assumption. However, this assumption breaks down in real-world scenarios, where previously unseen or unknown classes can appear during inference. Conventional models misclassify such unknown objects by assigning them to one or more known categories which can lead to incorrect scene understanding and pose significant safety risks.

To address this, research in this area has shifted towards anomaly segmentation \cite{Rba, Mask2Anomaly, Maskomaly} and open-world segmentation \cite{DML, ContMav}. Anomaly segmentation focuses on detecting and segmenting unknown objects absent during training, typically marking them as unknown without classifying the type of anomaly. open-world semantic segmentation goes beyond detecting unknowns: not only it segments known classes and flags novel regions, but also continuously incorporates newly discovered categories into the model via incremental learning, so that the recognized classes expand over time.
 Still, these approaches lack the ability to assign meaningful labels to unknown classes. These labels could not only help improve segmentation quality, but also help to distinguish different types of anomalies and support downstream tasks such as planning and control in autonomous driving. Works like AutoTAMP~\cite{chen2024autotamp} demonstrate that integrating vision-language aligned feature representations like CLIP~\cite{CLIP} into task and motion planning pipelines can improve planning robustness and generalization. This suggests that semantically meaningful representations, such as textual descriptions of objects or events, could potentially be leveraged in downstream planning tasks. While recent methods like ContMAV~\cite{ContMav} can distinguish between different unknown classes and apply continual learning techniques to adapt to them during training, they are still limited to predicting the closest known class and cannot provide semantically meaningful names for anomalies. Furthermore, the sparsity of training data for anomalies makes it difficult to distinguish between different anomalous objects that are similarly related to the known classes.

Recently, there has been increasing interest in using CLIP~\cite{CLIP} for open-vocabulary segmentation. Open-vocabulary segmentation allows models to generalize to novel classes using natural language guidance, typically leveraging vision-language models like CLIP. Despite their promise, open-vocabulary methods~\cite{CAT-Seg, SAN} still rely on a manually defined inference vocabulary, which cannot realistically cover all possible anomalies. Including too many labels can hurt performance on known classes and introduce ambiguity \cite{CAT-Seg}. As a result, these methods cannot be directly applied to anomaly segmentation. In practical autonomous driving scenarios, we require a system that segments and identifies anomalies using only a vocabulary of common known classes, while gracefully handling anything beyond that scope. 
An alternative direction is Vocabulary-Free Semantic Segmentation~\cite{AutoSeg, CASED, ZeroSeg}, which aims to segment an image without relying on a predefined set of class names at inference time. While recent methods in this category show promise, their performance still lags behind both open-vocabulary and closed-world models. 

We therefore propose a novel CLIP-based method that identifies anomalies with respect to the known vocabulary, assigns descriptive names to newly discovered anomalies, and continuously integrates them into the set of known classes.
%
Our main contributions can be summarized as:
\begin{itemize}
    \item A threshold-free novel anomaly segmentation approach based on CLIP that achieves state-of-the-art performance on benchmarks, significantly outperforming prior methods in mIoU.
    \item A method to assign meaningful labels to anomaly instances, outperforming SOTA open-world segmentation methods. 
    \item A way to integrate unknown classes into the known vocabulary at inference time, eliminating the need for continuous training required by previous open-world methods.

\end{itemize}

Our experiments show that the proposed method achieves state-of-the-art anomaly segmentation results, outperforms prior open-world approaches, preserves accuracy on known classes, and assigns meaningful semantic labels to anomalies.

\section{RELATED WORK}
\label{sec:related}
\subsection{Anomaly / Open-Set Segmentation}
Anomaly segmentation in autonomous driving addresses the challenge of detecting pixels that do not belong to any known class, often corresponding to out-of-distribution (OOD) objects, which is critical for safety.
SML \cite{SML} identifies anomalies by flagging pixels with unusually low standardized logits from a segmentation network. DenseHybrid \cite{DenseHybrid} improves on this by combining generative modeling with a discriminative branch, enhancing robustness in complex scenes. FlowCLAS~\cite{FlowCLAS} uses DINO~\cite{DINO} visual embeddings and a normalizing flow network to compute per-pixel likelihoods, optionally enhanced with contrastive learning. More recent mask-based approaches include Mask2Anomaly~\cite{Mask2Anomaly}, which leverages Mask2Former~\cite{Mask2Former} with global mask attention and contrastive learning to robustly separate anomalies from known classes, and Maskomaly \cite{Maskomaly}, which also uses Mask2Former but relies on softmax logits for simpler anomaly detection. Finally, UNO \cite{UNO} decouples negative samples from the no-object class, proposing a k+2 classifier for unsupervised novel object detection. 

\subsection{CLIP for Anomaly Detection}
While leveraging vision-language models such as CLIP for anomaly segmentation in autonomous driving has not been explored so far, similar approaches have been applied in other domains. WinCLIP~\cite{WinCLIP} leverages CLIP for zero- and few-shot anomaly classification and segmentation in industrial applications, identifying anomalous regions in products without task-specific training. Elhafsi et al.~\cite{Elhafsi2023} extend this idea to autonomous driving by using large language models to detect anomalous objects as bounding boxes in a zero-shot manner. He et al. \cite{SniffingOpenWorld} finetune an open-vocabulary method with a vocabulary consisting of the cityscapes classes and a general vocabulary for anomalies relevant to autonomous driving like "obstacle" and "roadblock" for open-world detection. 

\subsection{Open World Semantic Segmentation}

Open World semantic segmentation extends anomaly segmentation by not only detecting unknown regions but also incrementally incorporating them into the set of known classes, enabling the model to continuously learn and recognize novel categories over time. Cen et al. \cite{DML} proposed one of the first approaches for open-world semantic segmentation using deep metric learning and contrastive clustering to detect and incrementally incorporate novel classes. It requires OOD data for incremental learning of unknown classes. ContMAV~\cite{ContMav} introduces a contrastive loss and uses contrastive prototypes which naturally encode a distance from each unknown class to the known classes. This enables distinguishing different unknowns and allows the model to continuously learn representations of novel classes without requiring labeled out-of-distribution (OOD) data. 

Unlike ContMAV~\cite{ContMav}, which requires multiple observations to form a prototype for a novel class, we propose a method that can segment unknown classes in a zero-shot manner using pretrained vision-language embeddings and immediately integrate them into the known vocabulary. As a by-product, 
we can directly assign an accurate class name describing the anomaly. 

 \subsection{Open-Vocabulary \& Vocabulary-Free Segmentation}
 Zero-shot semantic segmentation of unknown classes using vision-language embeddings has been extensively studied under the paradigm of open-vocabulary semantic segmentation. Methods such as OV-Seg~\cite{OV-Seg}, SAN~\cite{SAN}, and CAT-Seg~\cite{CAT-Seg} demonstrate strong performance in segmenting novel classes by leveraging pretrained vision-language models. However, these approaches rely on a user-specified vocabulary at inference time, which limits their applicability in open-world scenarios where unexpected objects may appear. To overcome this, Vocabulary-Free Semantic Segmentation has been proposed, aiming to operate without predefined vocabularies. Approaches like CaSED~\cite{CASED} and Chicken-and-Egg~\cite{OVVF} show that it is possible to discover novel categories automatically, but current results still fall short of the accuracy achieved by open-vocabulary methods. This is why we propose a novel combined approach.

\begin{figure*}[t]
    \centering
    \includegraphics[width=\textwidth]
    {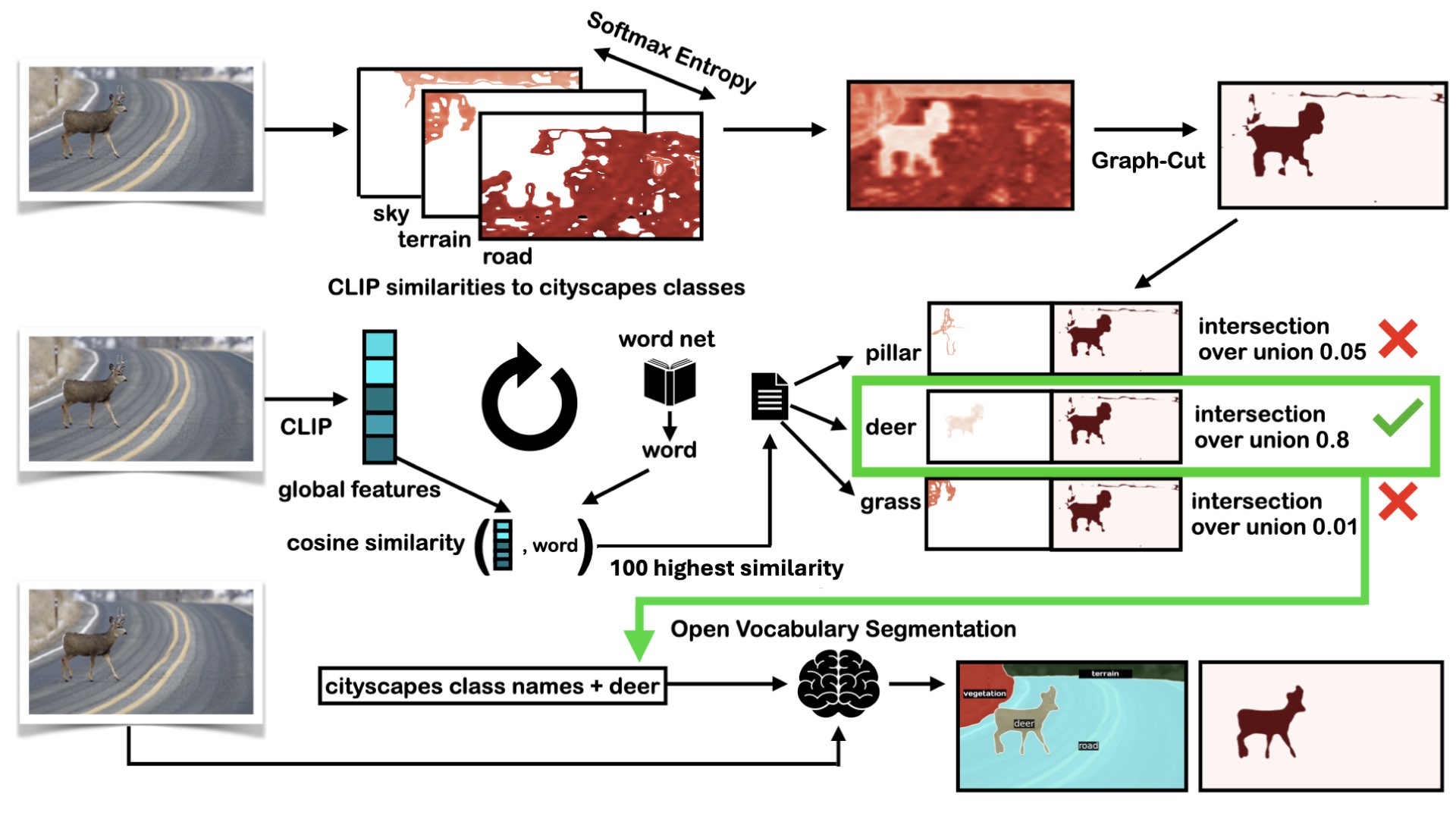}
    \caption{Overview of our Method Clipomaly: For clarity this figure only shows our dictionary-based method together with \textsc{CLIP-Best} Region Matching.} 
    \label{fig:method}
\end{figure*}
\section{METHOD}
\label{ch:method}


We formulate the problem of open-world semantic segmentation in terms of known and unknown vocabularies. We assume that we are given a known vocabulary and need to generate a vocabulary for the unknown classes. 
Given an input image and a known vocabulary $\mathcal{V} = \{v_1, v_2, \ldots, v_C\}$ available at inference time, the goal is to produce a segmentation map over an extended vocabulary $\mathcal{V} \cup \mathcal{W}$, where $\mathcal{W} = \{w_1, w_2, \ldots, w_k\}$ is a set of previously unseen or unknown semantic concepts not included in $\mathcal{V}$.
Note that, where not mentioned otherwise, our known vocabulary $\mathcal{V}$ is the class names of the Cityscapes dataset~\cite{Cityscapes}. 
Our method to solve this problem consists of the following steps: 
\begin{itemize}
    \item \emph{Unknown Mask Prediction}: Identify regions in the image that do not correspond to any $v_i \in \mathcal{V}$ (i.e., \textit{unknown} regions);
    \item \emph{Tagging}: Generate a new set of semantic labels $\mathcal{W} = \{w_1, w_2, \ldots, w_k\}$ for these unknown regions, where $\mathcal{W} \cap \mathcal{V} = \emptyset$;
    \item \emph{open-vocabulary segmentation}: Determine a final open-vocabulary segmentation over the extended vocabulary $\mathcal{V} \cup \mathcal{W}$.
\end{itemize}




\subsection{Unknown Mask Prediction}
\label{sec:unknown_mask_prediction}

We extract dense CLIP \cite{CLIP} embeddings from the last attention layer, following the approach introduced in \cite{CAT-Seg}. To enable open-vocabulary segmentation, we use a CLIP model fine-tuned on semantic segmentation on our known vocabulary $\mathcal{V}$. 
We observe that dense CLIP embeddings, even after fine-tuning, retain the ability to identify regions that do not correspond to any class in the known vocabulary. 
We define the following:
\begin{itemize}
    \item $C = |\mathcal{V}|$, the number of classes
    \item $E_{i,j} \in \mathbb{R}^d$, the dense CLIP image embedding at spatial location \( (i,j) \)
    \item $t_v^c \in \mathbb{R}^d$, the text embedding of class $v_c \in \mathcal{V}$
\end{itemize}

For a given image, we obtain a dense similarity map between dense CLIP features and the text embeddings, allowing us to estimate class-wise relevance across the scene.
\[
S \in \mathbb{R}^{H \times W \times C}, \quad
S_{i,j}^c = \cos\left( E_{i,j}, \, t_v^c \right) = \frac{E_{i,j} \cdot t_v^c}{\|E_{i,j}\| \, \|t_v^c\|}
\]

To quantify uncertainty at the pixel level, we compute the \textit{pixelwise softmax entropy} over the class dimension:
\[
\text{Entropy}(i, j) = - \sum_{c=1}^C p_{i,j}^{(c)} \log p_{i,j}^{(c)}
\]
\[
\text{where} \quad p_{i,j}^{(c)} = \frac{\exp(S_{i,j}^{(c)})}{\sum_{c'=1}^C \exp(S_{i,j}^{(c')})}.
\]


To improve the usability of the entropy map for Graph Cut \cite{GraphCut}, we normalize it to $[0, 1]$. 
\[
\hat{E}(i, j) = \frac{\text{Entropy}(i, j) - \min(\text{Entropy})}{\operatorname{max}(\text{Entropy}) - \operatorname{min}(\text{Entropy})}.
\]

This normalized uncertainty map $\hat{E}$ enhances the contrast between confident and ambiguous regions, making it easier to extract spatially coherent unknown areas. We apply \textit{Graph Cut}~\cite{GraphCut} to this map to obtain the final segmentation of unknown regions
. For Graph Cut, the bottom 10\% of normalized uncertainty pixels (most confident) are source seeds, and the top 10\% (most uncertain) are target seeds. 
 



\subsection{Tagging}
\label{sec:unknown}

We generate semantic labels in two steps. 

\begin{itemize}
\item Generate a low number of meaningful text candidates.
\item Choose the text candidates that most closely match with the produced unknown mask and represent the unknown vocabulary $\mathcal{W}$.
\end{itemize}




\subsubsection{Candidate Text Generation} 
We explore two strategies. 
One is more lightweight and uses a dictionary to preselect meaningful candidates. The other one uses an additional image tagging model. Specifically, we use Recognize-Anything Model (RAM)~\cite{RAM}, an open-vocabulary image tagging model capable of predicting a wide range of object-related terms. Due to its ability to generate semantically rich and highly relevant labels for most visible objects, RAM yields strong performance in our setting. However, it is computationally expensive and may not be practical for deployment in resource-constrained environments such as autonomous vehicles. 

On the other hand, our dictionary based strategy is more lightweight and therefore an effective alternative. We employ a large, fixed vocabulary 
\mbox{$\mathcal{D}=\{d_1,d_2,\ldots,d_N\}$} 
of candidate words. We use a preselection mechanism based on aggregated similarity to identify a small subset of promising candidates from \(\mathcal{D}\), reducing computational cost while maintaining reasonable quality.
 We obtain the global image embedding \( \mathbf{e}_{\text{img}} \in \mathbb{R}^d \) from the CLIP image encoder. For each candidate text \( t_k \) in the dictionary \( \mathcal{D} \), we then compute its text embedding \( \mathbf{e}_{t_k} \in \mathbb{R}^d \) and measure the cosine similarity with the image embedding. 


\[
\text{sim}(t_k) = \frac{\langle \mathbf{e}_{\text{img}}, \mathbf{e}_{t_k} \rangle}{\|\mathbf{e}_{\text{img}}\| \cdot \|\mathbf{e}_{t_k}\|}
\]

We select the top-\(K\) candidates with the highest similarity scores. These candidates are then used in the next stage to assign semantic labels to unknown regions.
A similar preselection step is used in CaSED \cite{CASED}, a vocabulary-free semantic segmentation method. In both methods, this preselection substantially reduces computation compared to dense or patchwise similarity, making it well-suited for real-time and embedded applications. 

\subsubsection{Candidate Selection via Region Matching}

To identify the best-fitting candidate from the generated list, we compute a spatial similarity based on dense CLIP embeddings. This enables us to directly match segmentation masks to words using dense CLIP representations. In contrast to OV-Seg \cite{OV-Seg}, which relies on a fine-tuned CLIP variant that accepts masks as input, our approach leverages the original CLIP model without the need for additional fine-tuning on masks.

In detail, we compute an unknown mask for each candidate text \(t_k\) as follows.
For each spatial location \((i, j)\), we determine the class from the known vocabulary $\mathcal{V}$ with the highest similarity score:
\[
c_{i,j} = \arg\max_{c \in \mathcal{V} \cup \{t_k\}} S_{i,j}^{(c)}
\]
where \(S_{i,j}^{(c)}\) denotes the similarity between the dense embedding at \((i, j)\) and the text embedding of class \(c\).


We define the soft mask \(M_{t_k}\) for text candidate \(t_k\) as:
\[
M_{t_k}(i, j) =
\begin{cases}
\max_c S_{i,j}^{(c)} & \text{if } c_{i,j} = t_k \\
0 & \text{otherwise}
\end{cases}
\]

Let \( M_{\text{unk}} \) be the binary mask of the unknown region, computed as described in Section \ref{sec:unknown_mask_prediction}.
We compute the soft Intersection-over-Union (IoU) between \( M_{t_k} \) and \( M_{\text{unk}} \) as:
\[
\frac{\sum_{i,j} M_{t_k}(i, j) \cdot M_{\text{unk}}(i, j)}%
{\sum_{i,j} \left[ M_{t_k}(i, j) + M_{\text{unk}}(i, j) - M_{t_k}(i, j) \cdot M_{\text{unk}}(i, j) \right]}
\]


Finally, we select
$ t^* = \arg\max_{t_k} \text{IoU}(t_k)$
and as unknown vocabulary $\mathcal{W} = \{t^*\}$. For the RAM-based tagging variant, we found it more effective to bypass the soft IoU computation and instead rank candidate texts by averaging the similarity values within the region mask $M_{t_k}$ for each text $t_k$. This is because RAM typically produces high-quality candidate labels, making the additional graph-cut-based soft IoU step unnecessary.

We call this strategy \textsc{CLIP-BEST}, as it ensures that the selected text most closely matches the spatial layout of the unknown region. 
As an alternative we also explore \textsc{CLIP-ALL}, an iterative variant: we first select the candidate with the highest soft IoU, integrate it into the known vocabulary, and then recompute the soft IoU of the unknown region with the remaining candidates. Additional candidates are added one by one as long as the combined soft IoU improves, stopping once no further improvement is possible.

\subsection{Open-Vocabulary Segmentation}

For open-vocabulary segmentation, we adopt CAT-Seg \cite{CAT-Seg} as our segmentation backbone due to its high performance. We supply the model with inference class names  
$\mathcal{V} \cup \mathcal{W}$, with known vocabulary $\mathcal{V}$ and unknown vocabulary $\mathcal{W}$ obtained as described in Section \ref{sec:unknown}.

This extended vocabulary enables the model to assign semantically meaningful labels to both known and previously unlabeled regions. Specifically, the model produces a segmentation map over $\mathcal{V} \cup \mathcal{W}$. To distinguish between known and unknown regions, we post-process the predicted segmentation mask: all pixels predicted as one of the known texts in $ \mathcal{V}$ are labeled as known, while those predicted as unknown texts in $\mathcal{W}$ are labeled as unknown. This yields both a fine-grained open-vocabulary segmentation and a binary known/unknown segmentation.

\section{EXPERIMENTS}
\label{ch:results}

In this section, we demonstrate the effectiveness of our approach across several challenging benchmarks and tasks. 
We highlight three main aspects: 
\begin{itemize}
    \item Our method achieves state-of-the-art performance on anomaly segmentation datasets such as SMIYC~\cite{SMIYC} and RoadAnomaly~\cite{RoadAnomaly}.
    \item We outperform the leading open-world segmentation method ContMAV~\cite{ContMav} in anomaly and open-world segmentation settings.
    \item Our model produces human-understandable descriptions of anomalies while simultaneously delivering high-quality open-vocabulary segmentations, showing its versatility within a single framework.
\end{itemize}

\subsection{Experimental Setting}
\textbf{Datasets}: RoadAnomaly~\cite{RoadAnomaly} consists of 60 images with ground truth anomaly segmentations. The images contain obstacles like lost tires or stones and anomalies like animals (30 images) and weird looking vehicles. SMIYC~\cite{SMIYC} AnomalyTrack is a public benchmark. It consists in total of 110 images. Among them, 10 images are provided with ground truth anomaly segmentations and are used as the validation dataset. The remaining 100 images function as the test dataset and are published without ground truth anomaly segmentations. The metrics on the test dataset can only be computed by submitting to the public leaderboard. 

\textbf{Metrics}: Following common anomaly segmentation practice, we report Area Under the Precision Recall Curve (AUPR), False Positive Rate at 95\% True Positive Rate (FPR95), and False Positive Rate at 90\% True Positive Rate (FPR90) to account for threshold dependence. Anomaly segmentation performance is measured with Mean Intersection over Union (mIoU), or soft mIoU (smIoU) for methods producing heat maps. For validation datasets, we also report the Area under the Receiver Operator Curve (AuROC).

\textbf{Hyperparameters}: Unlike other anomaly segmentation methods we do not need a threshold to generate a segmentation from a produced anomaly heatmap. We generate a segmentation and a probability per pixel naturally from open vocabulary semantic segmentation.
We use CLIP~\cite{CLIP} with a  Vit-B/16 backbone, finetuned on Cityscapes~\cite{Cityscapes} for open vocabulary segmentation with CAT-Seg~\cite{CAT-Seg}. The training parameters follow \cite{CAT-Seg}. The batch size is 4, and models are trained for 80k iterations.

\subsection{Anomaly Segmentation}

We compare our method against anomaly segmentation approaches that do not use explicit OOD data. Although our approach leverages CLIP, which has been pretrained on web-scale data containing diverse concepts, we do not provide the model with any images explicitly labeled as anomalies or OOD during training. In other words, we do not use additional OOD supervision. The model only learns from standard in-distribution data. Any OOD knowledge comes indirectly from the pretrained representation which was not trained on segmentation. 

Table \ref{tab:roadanomaly_benchmark} shows that we perform best in AUPR and mean intersection over union with our RAM based approach. Only in FPR95 we do not perform best. 
However, in our method false positives are less critical in practice. False positives refer to in-vocabulary pixels being predicted as anomaly. In our method these pixels still receive a semantically meaningful label from the open vocabulary segmentation. 
Our more lightweight dictionary method performs better than current state-of-the-art methods in mIoU as well which shows our superior segmentation performance. In terms of AUPR, our dictionary-based approaches perform worse than UNO~\cite{UNO}, but still outperform several other state-of-the-art methods. Table \ref{tab:smiyc_benchmark} reports our results on the SMIYC public leaderboard, where we rank sixth in AUPR and ahead of ContMAV~\cite{ContMav}, the highest-scoring method that also provides both anomaly and open-world segmentation.

Table \ref{tab:ours_better_cityscapes} shows that we outperform Maskomaly~\cite{Maskomaly} in mIoU on the SMIYC validation set. The mean intersection over union cannot be reported on the public benchmark, as ground-truth labels are not available. Table \ref{tab:ours_better_cityscapes} further indicates that our RAM-based method surpasses our dictionary-based approaches, although this advantage is less pronounced on the validation set, where more accurate initial mask predictions reduce the impact of false positives.

\begin{table}[h]
    \centering
    \caption{Comparison on RoadAnomaly. We additionally report mIoU to emphasize segmentation performance.}
    \label{tab:roadanomaly_benchmark}
    \begin{tabular}{l c ccc }
        \toprule
        \textbf{Method} & \begin{tabular}{@{}c@{}} No OoD \\ Data\end{tabular} & AUPR↑ & FPR95↓ & mIoU↑ \\ 
        \midrule
        Ours – RAM   & \ding{51} & \textbf{84.7} & 14.9 & \textbf{57.8} \\ 
        UNO~\cite{UNO}                 & \ding{51} & \underline{82.4} & \textbf{9.2}  & - \\   
        Ours – Dict  & \ding{51} & 76.5 & 27.4 & \underline{53.0} \\ 
        Maskomaly~\cite{Maskomaly}     & \ding{51} & 70.9 & \underline{11.9} & 45.0 \\  
        EAM~\cite{EAM}                 & \ding{51} & 66.7 & 13.4 & – \\ 
        GMMSeg~\cite{GMMSeg}           & \ding{51} & 57.7 & 44.3 & – \\
        ObsNet~\cite{ObsNet}           & \ding{51} & 54.7 & 60.0 & – \\
        DenseHybrid~\cite{DenseHybrid} & \ding{51} & 35.1 & 43.2 & – \\
        SML~\cite{SML}                 & \ding{51} & 25.8 & 49.7 & – \\
        ML~\cite{ML}                   & \ding{51} & 19.0 & 70.5 & – \\
        \bottomrule
    \end{tabular}
\end{table}

\begin{table}[h]
    \centering
    \caption{Comparison on AnomalyTrack of SMIYC public leaderboard.}
    \label{tab:smiyc_benchmark}
    \begin{tabular}{l c c c cc}
        \toprule
        \textbf{Method} & \begin{tabular}{@{}c@{}}No OoD \\ Data\end{tabular} & \begin{tabular}{@{}c@{}}OW \\ Seg.\end{tabular} & \begin{tabular}{@{}c@{}}OV \\ Seg.\end{tabular} & AUPR↑ & FPR95↓ \\
        \midrule
        UNO~\cite{UNO}             & \ding{51} & \ding{55} & \ding{55} & \textbf{96.10} & \textbf{2.27} \\
        SOTA-UEM~\cite{UEM}        & \ding{51} & \ding{55} & \ding{55} & \underline{95.77} & \underline{2.49} \\
        FlowCLAS~\cite{FlowCLAS}        & \ding{51} & \ding{55} & \ding{55} & 94.33 & 6.58 \\
        SOTA-RbA~\cite{Rba}        & \ding{51} & \ding{55} & \ding{55} & 93.89 & 7.86 \\
        Maskomaly~\cite{Maskomaly} & \ding{51} & \ding{55} & \ding{55} & 93.55 & 6.87 \\
        \midrule
        Ours - RAM & \ding{51} & \ding{51} & \ding{51} & 92.23 & 8.78 \\
        ContMAV~\cite{ContMav}         & \ding{51} & \ding{51} & \ding{55} & 90.20 & 3.83 \\
        \bottomrule
    \end{tabular}
\end{table}

\begin{table}[h]
    \centering
    \caption{Comparison of text generation strategies using mIoU on RoadAnomaly test set, tested on ground truth masks.}
    \label{tab:textgen_match}
    \begin{tabular}{l c c}
        \toprule
        \textbf{Method} & \textbf{CLIP Match} &   \textbf{RoadAnomaly} \\
        \midrule
        RAM & Best  & 63.5 \\
        RAM & All  & \underline{64.5} \\
        Dict & Best  & 59.3 \\
        Dict & All & \textbf{65.5} \\
        \bottomrule
    \end{tabular}
\end{table}

\begin{figure*}[ht]
    \centering
    \begin{tabular}{@{}cccc@{}}
        \textbf{Ground Truth} & \textbf{Ours CLIP-Best} & \textbf{Ours CLIP-All} & \textbf{Maskomaly} \\[0.2cm]
        
        \includegraphics[width=0.23\textwidth]{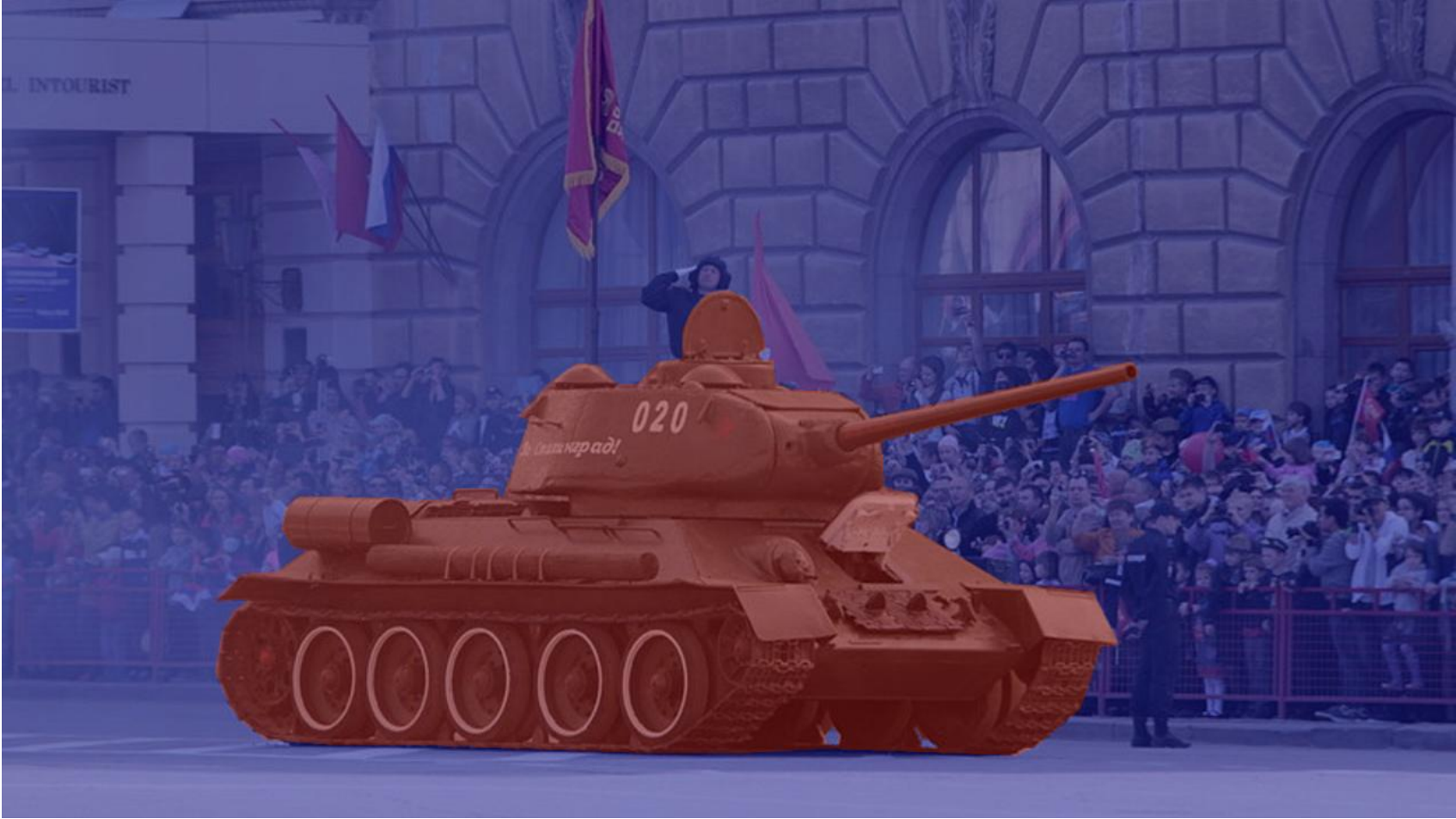} &
        \includegraphics[width=0.23\textwidth]{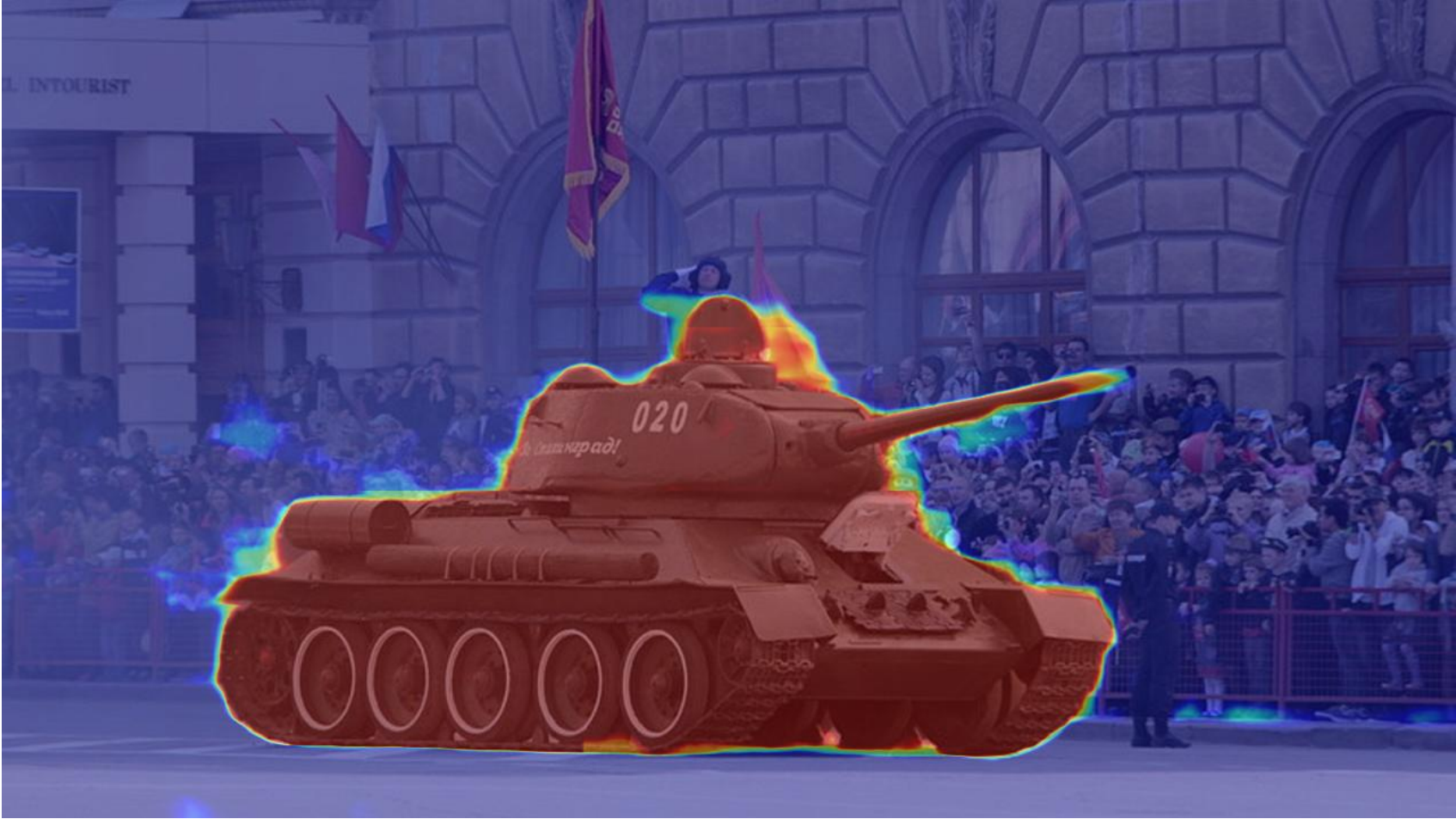} &
        \includegraphics[width=0.23\textwidth]{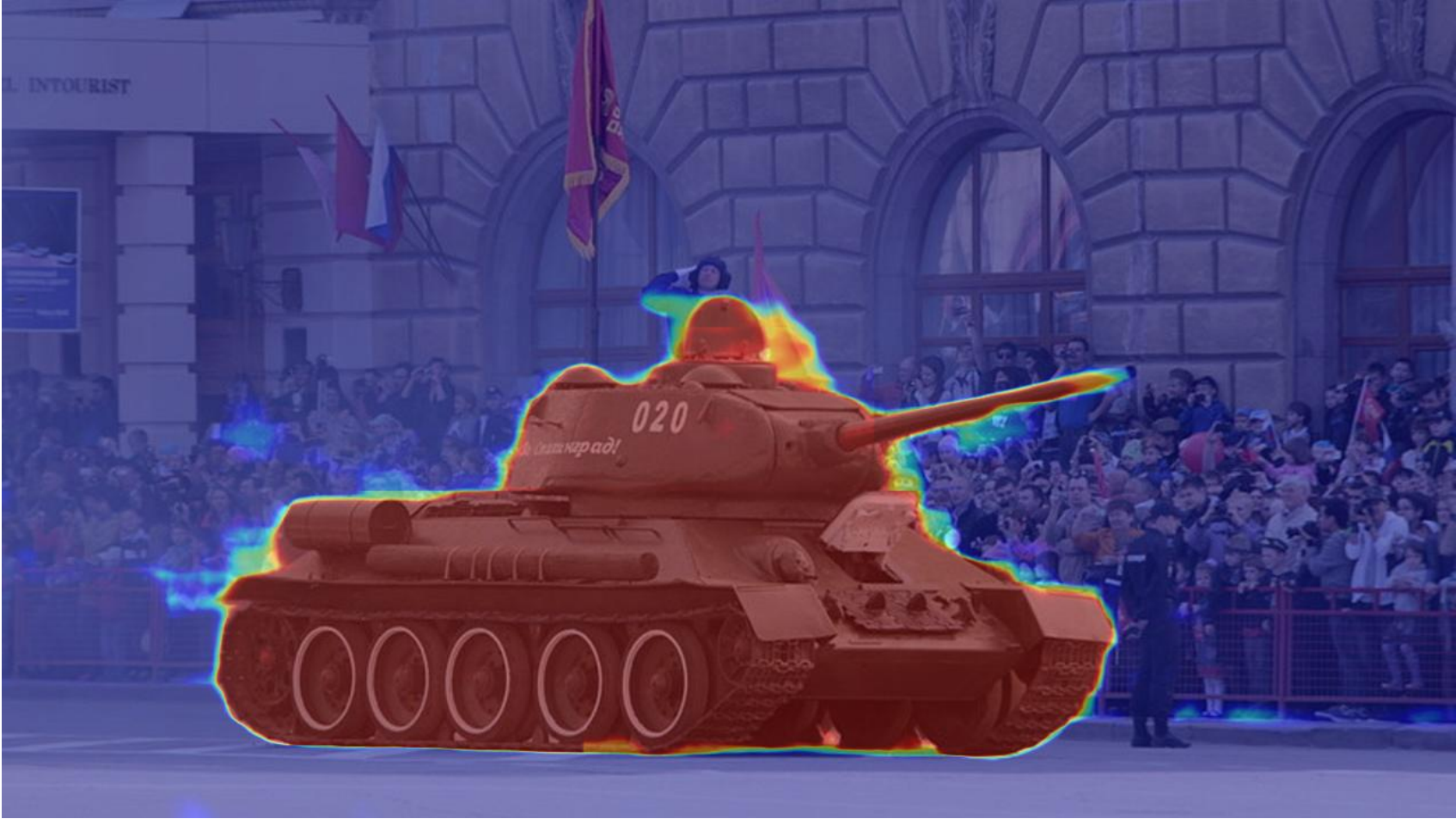} &
        \includegraphics[width=0.23\textwidth]{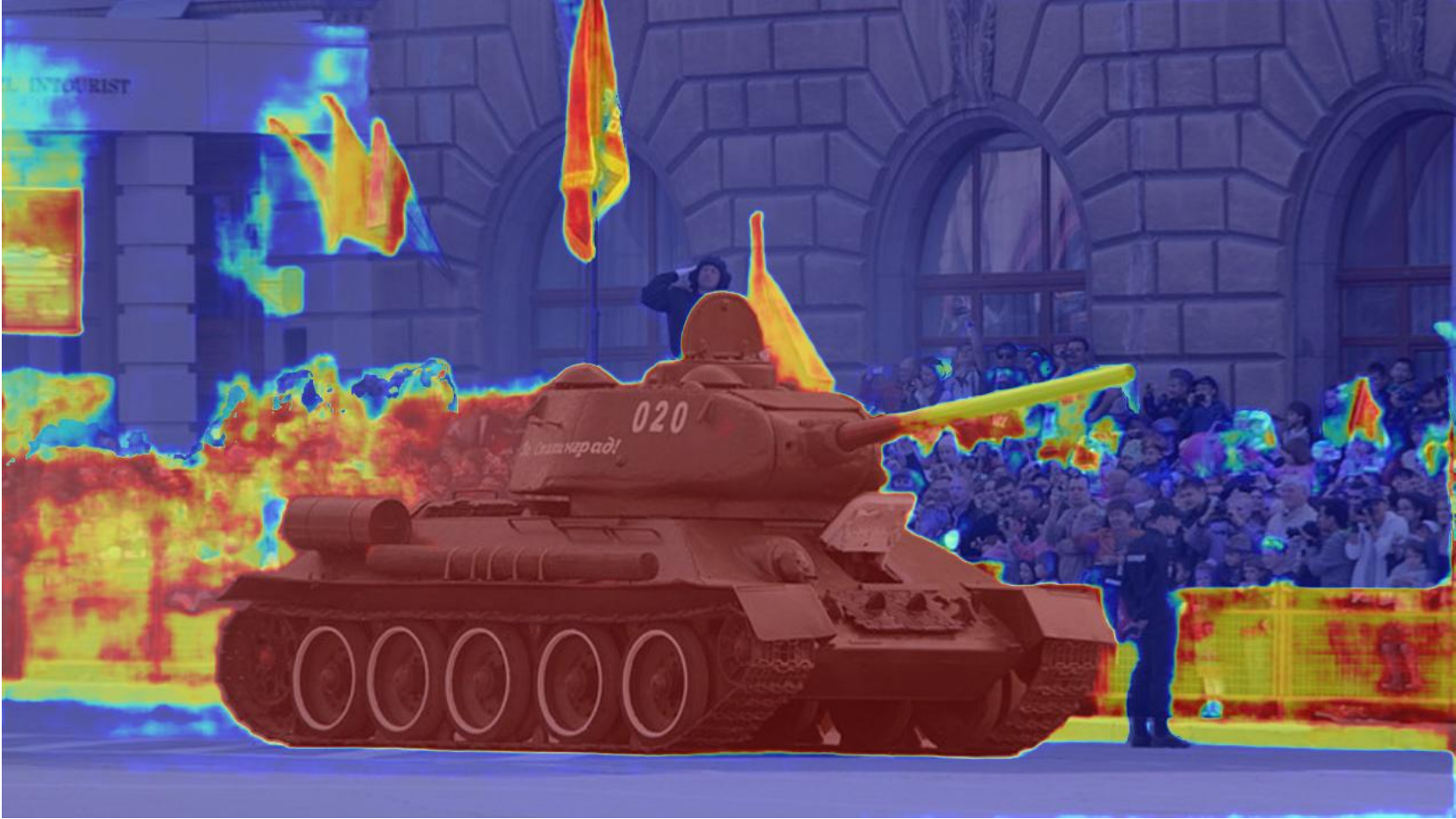} \\
        [-0.2cm]
         &
        \makebox[0.12\textwidth]{\scriptsize "army tank"} &
        \makebox[0.12\textwidth]{\scriptsize "army tank"} 
        \\ [0.1cm]
        
        \includegraphics[width=0.23\textwidth]{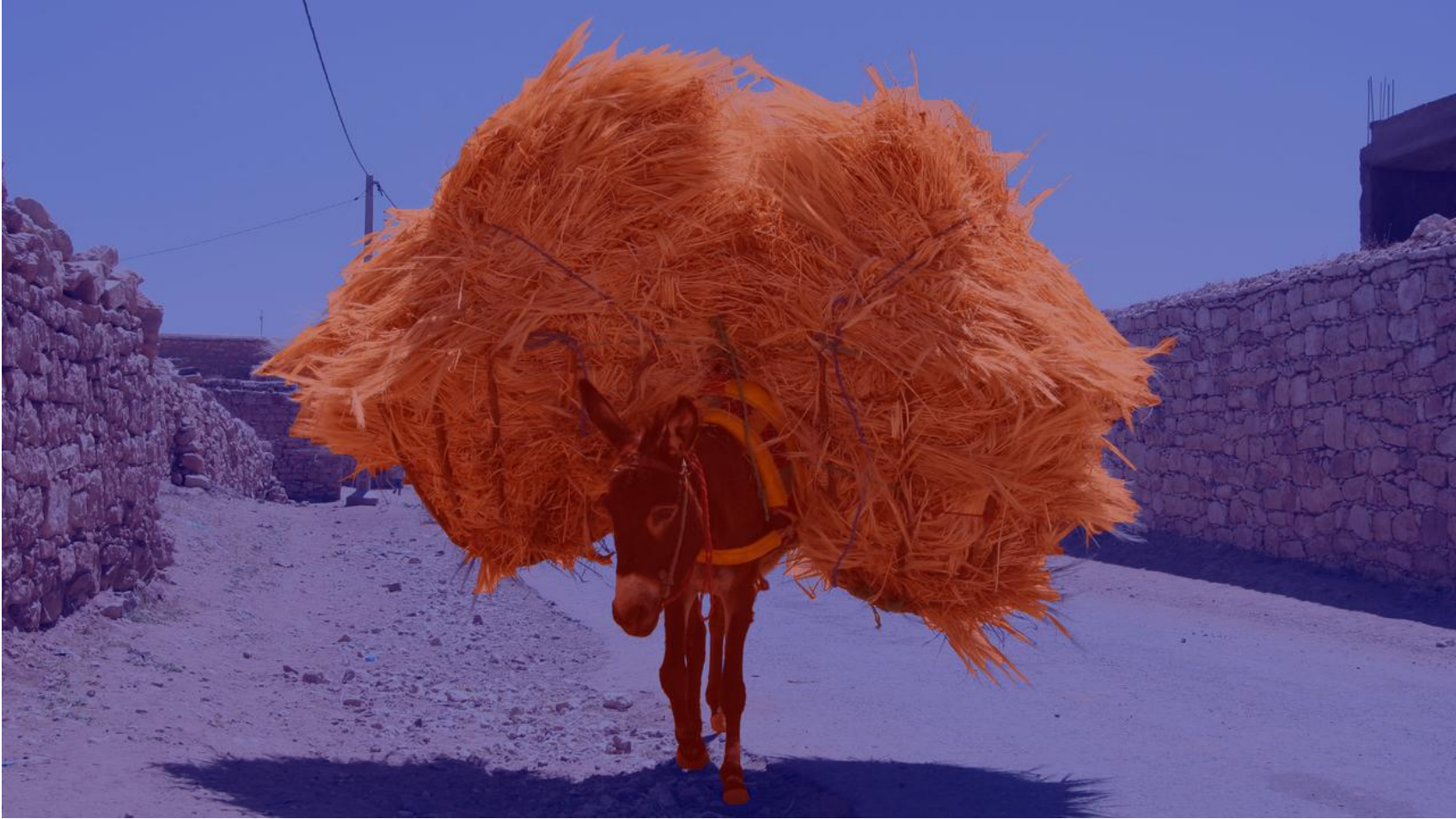} &
        \includegraphics[width=0.23\textwidth]{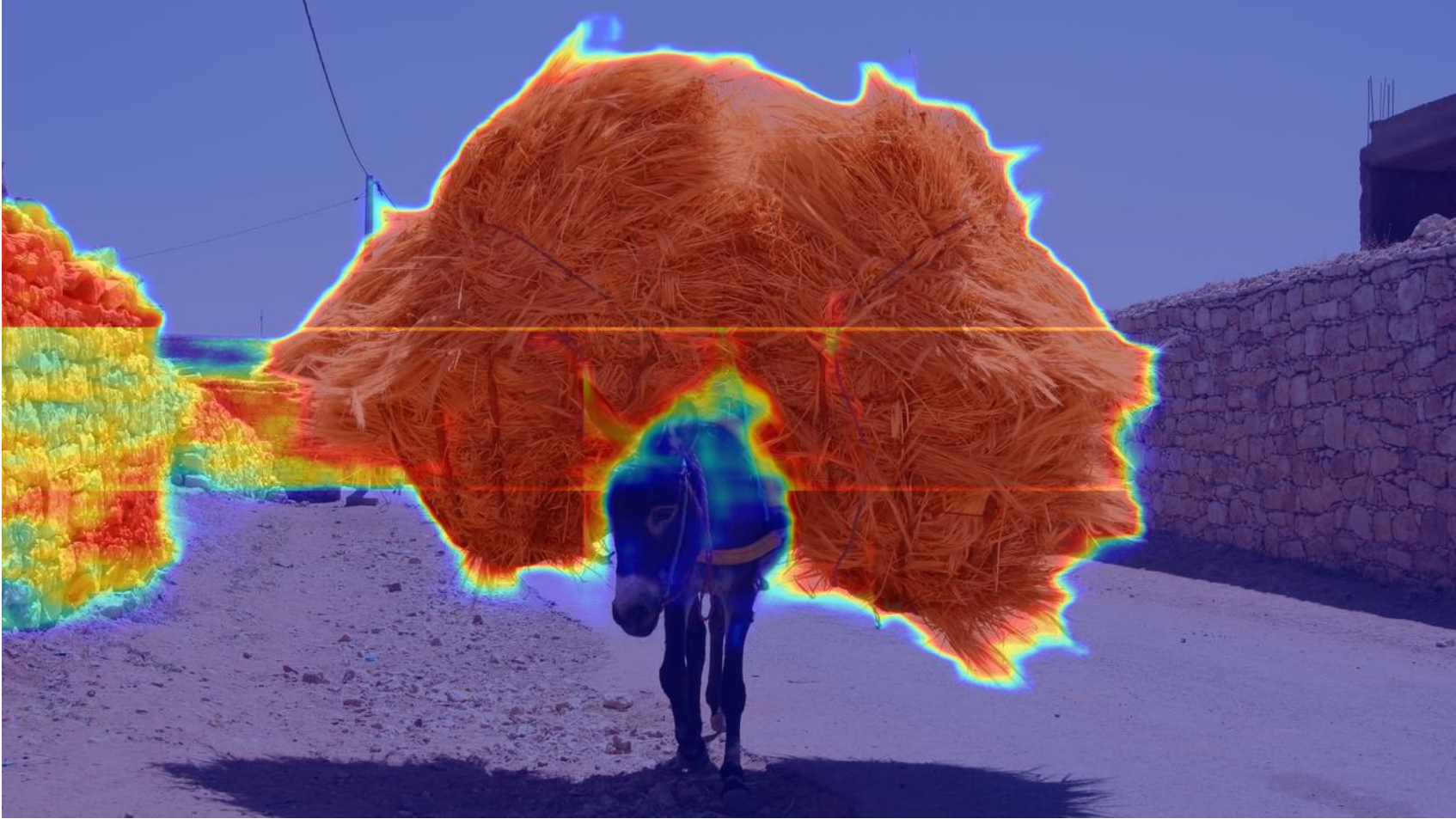} &
        \includegraphics[width=0.23\textwidth]{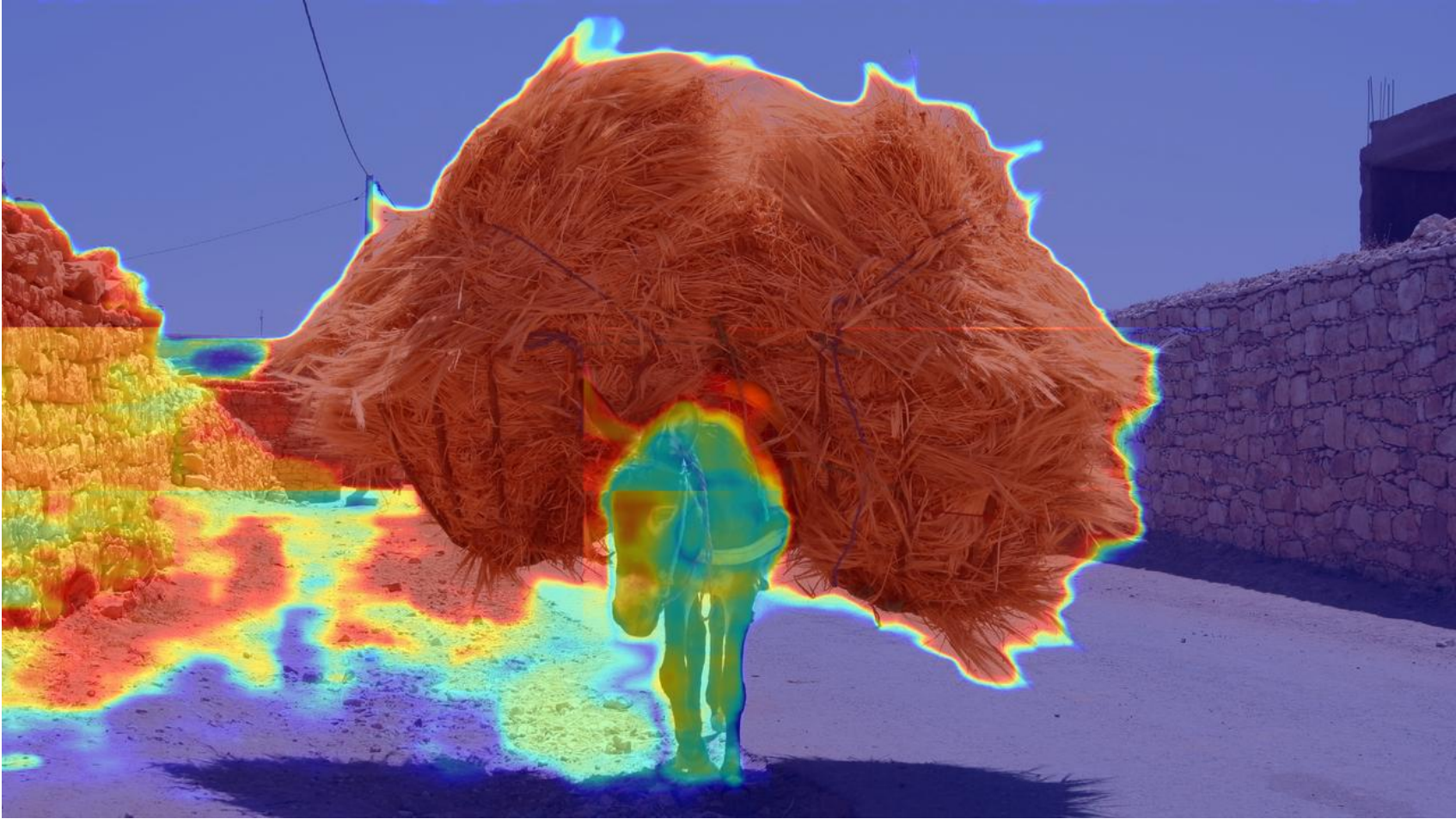} &
        \includegraphics[width=0.23\textwidth]{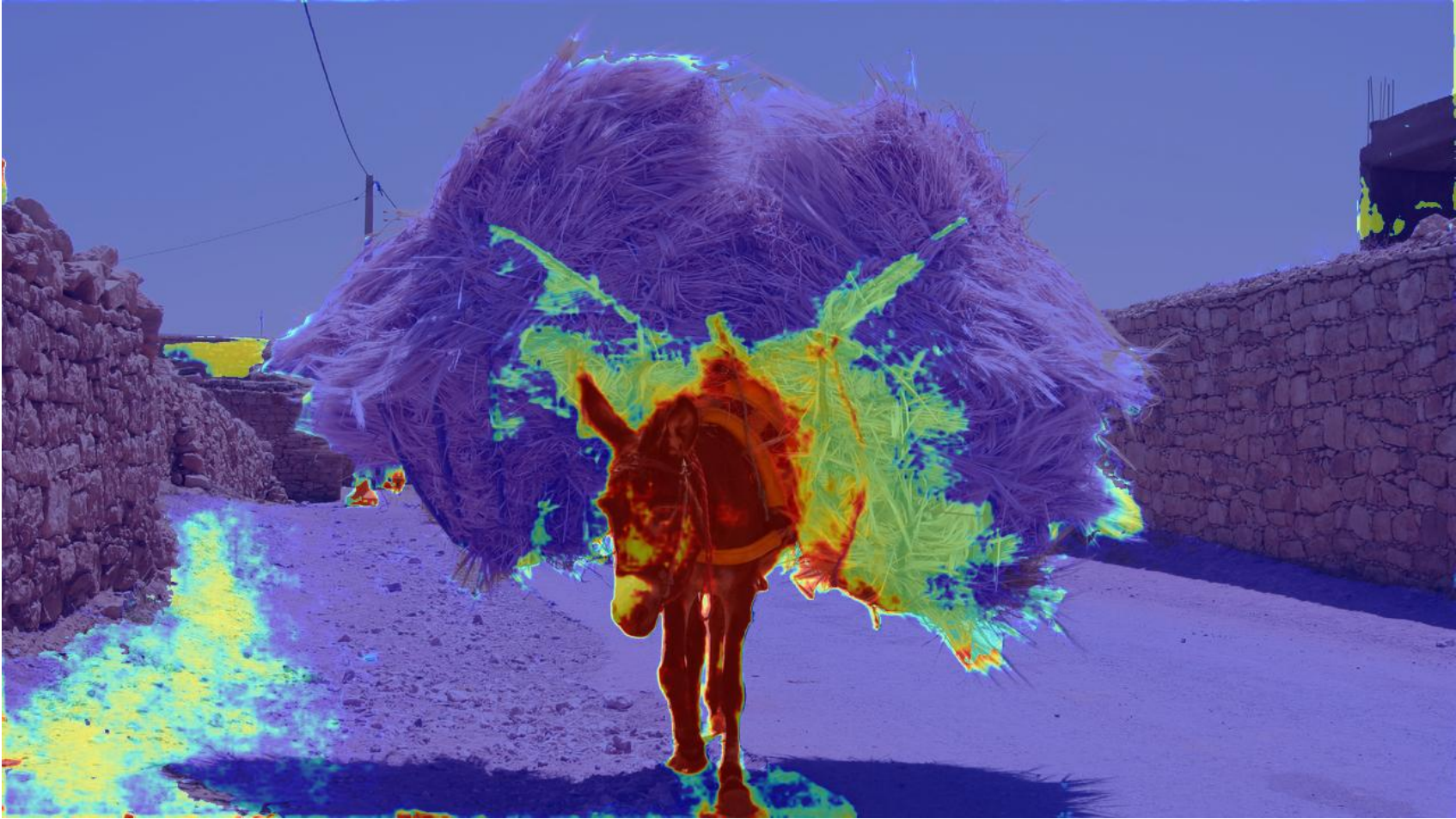} \\
        [-0.2cm]
        &
        \makebox[0.12\textwidth]{\scriptsize "hay"} &
        \makebox[0.12\textwidth]{\scriptsize "hay", "mule", "donkey"} \\ [0.1cm]

        \includegraphics[width=0.23\textwidth]{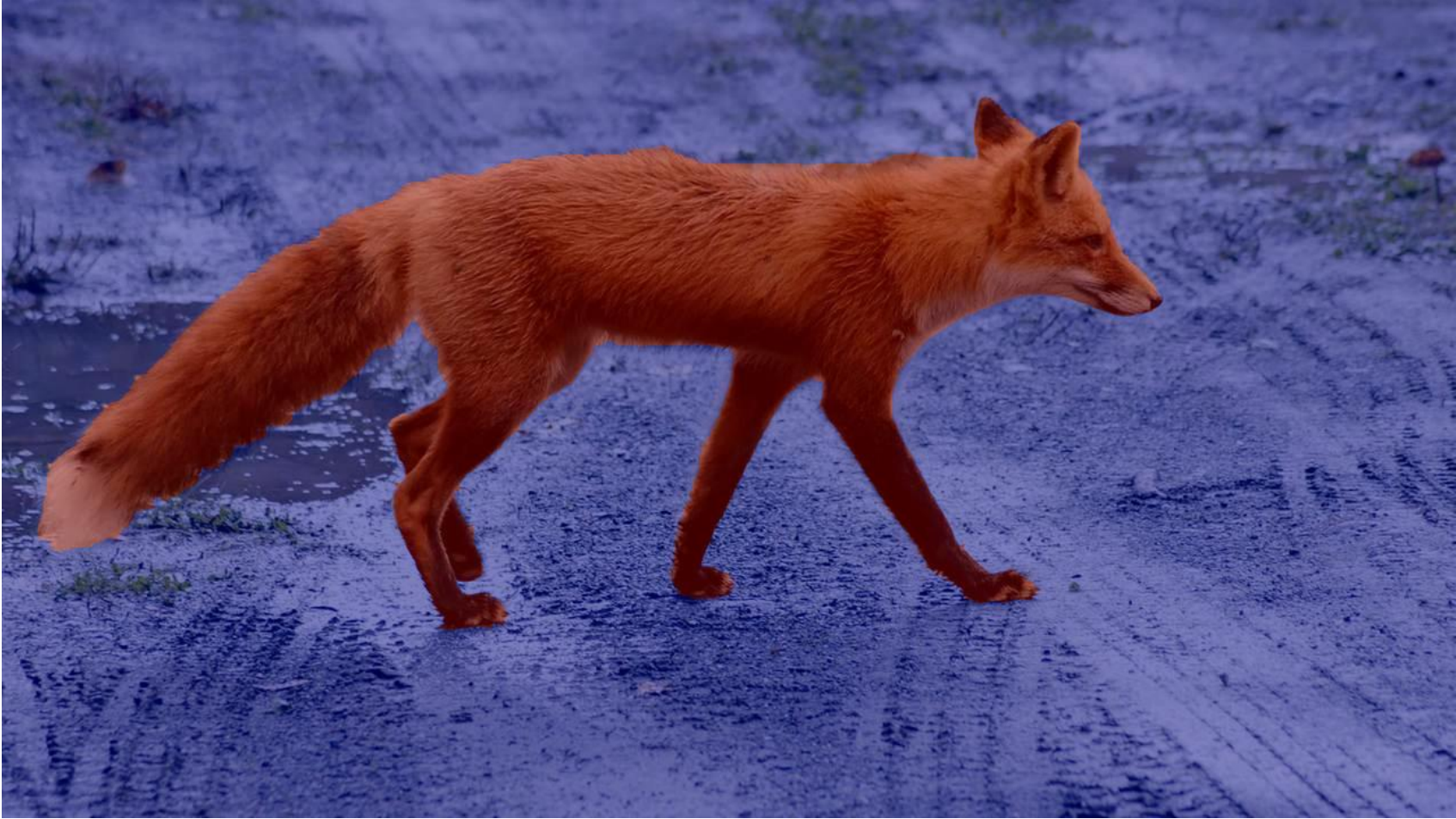} &
        \includegraphics[width=0.23\textwidth]{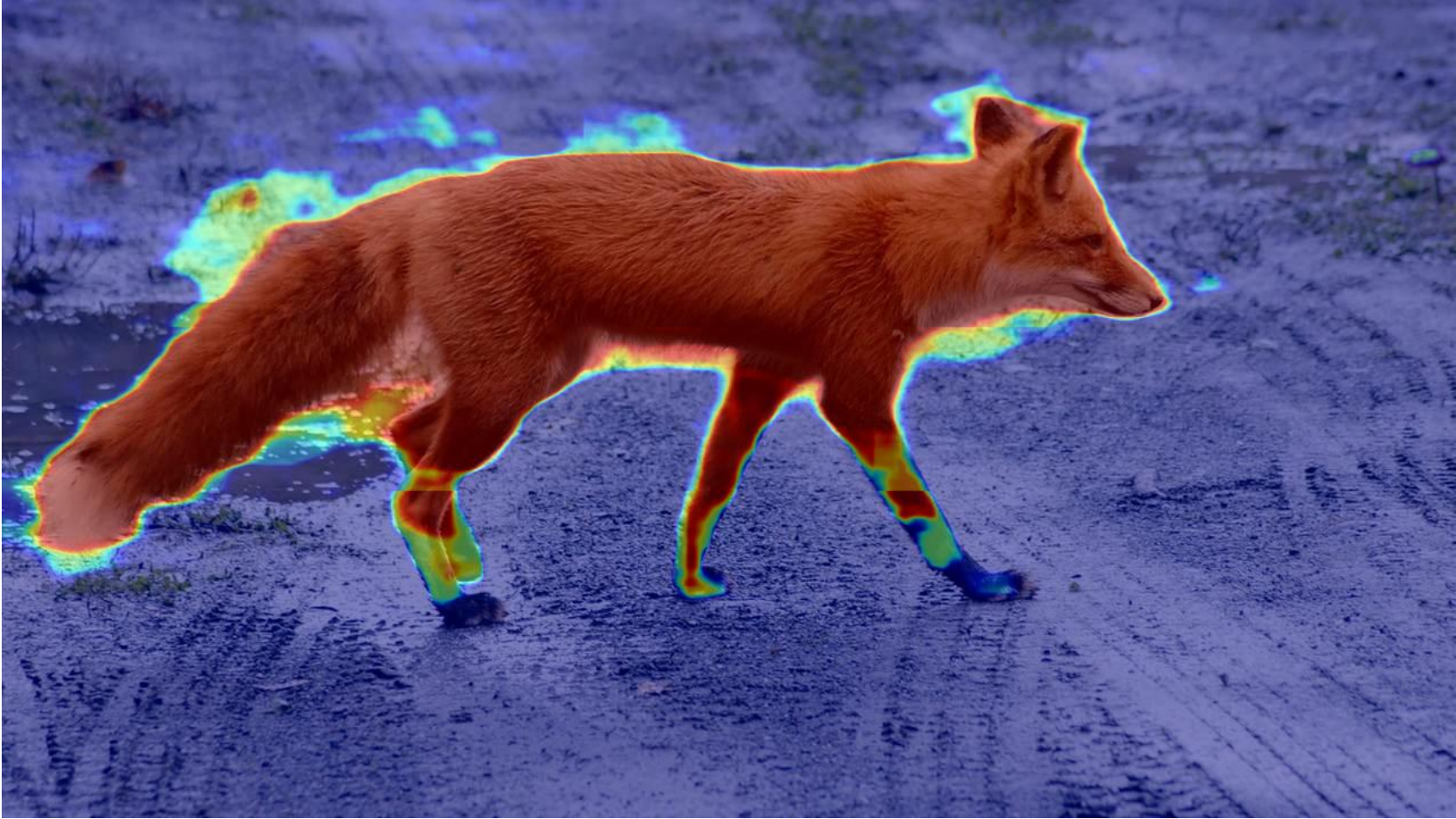} &
        \includegraphics[width=0.23\textwidth]{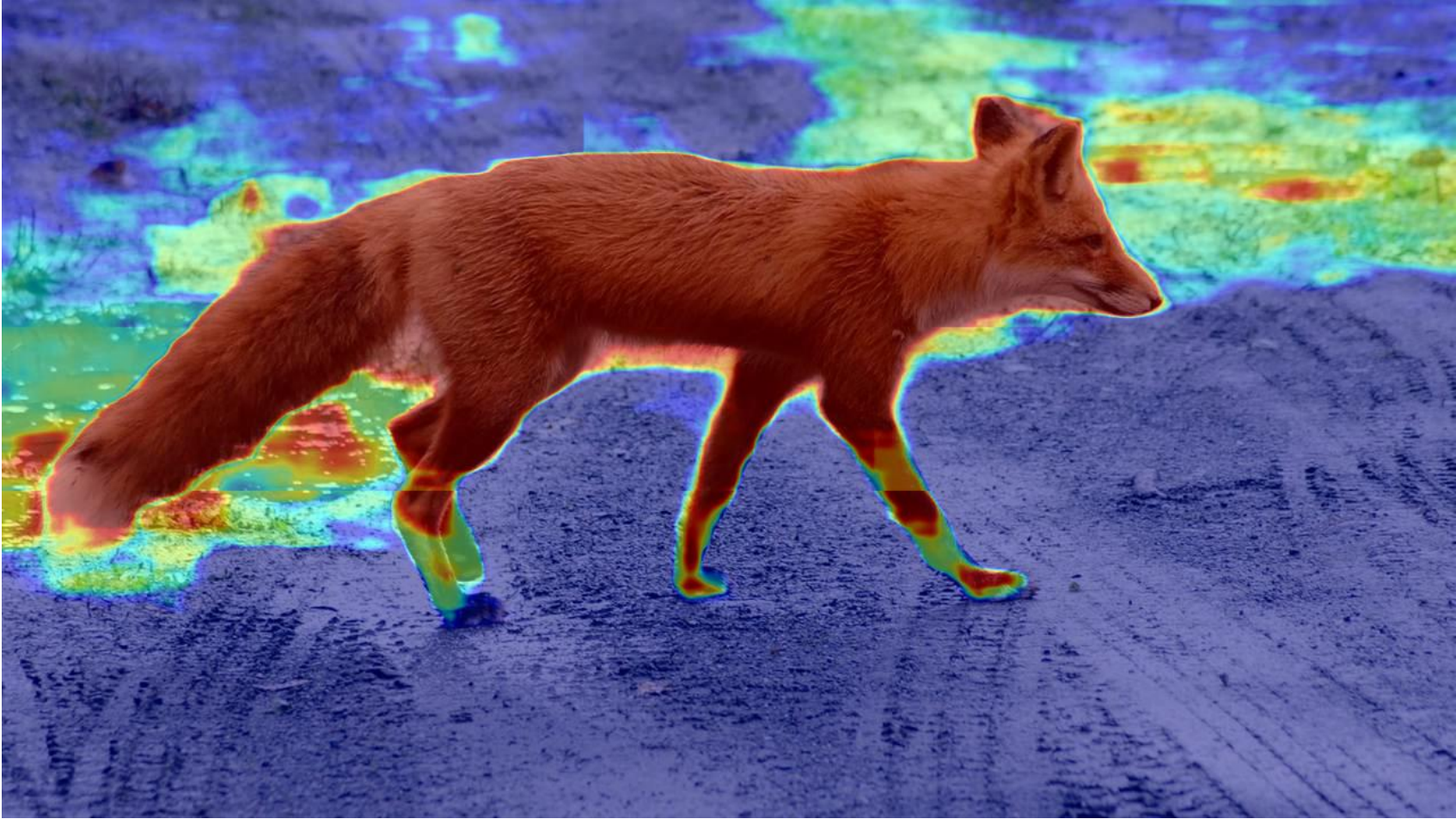} &
        \includegraphics[width=0.23\textwidth]{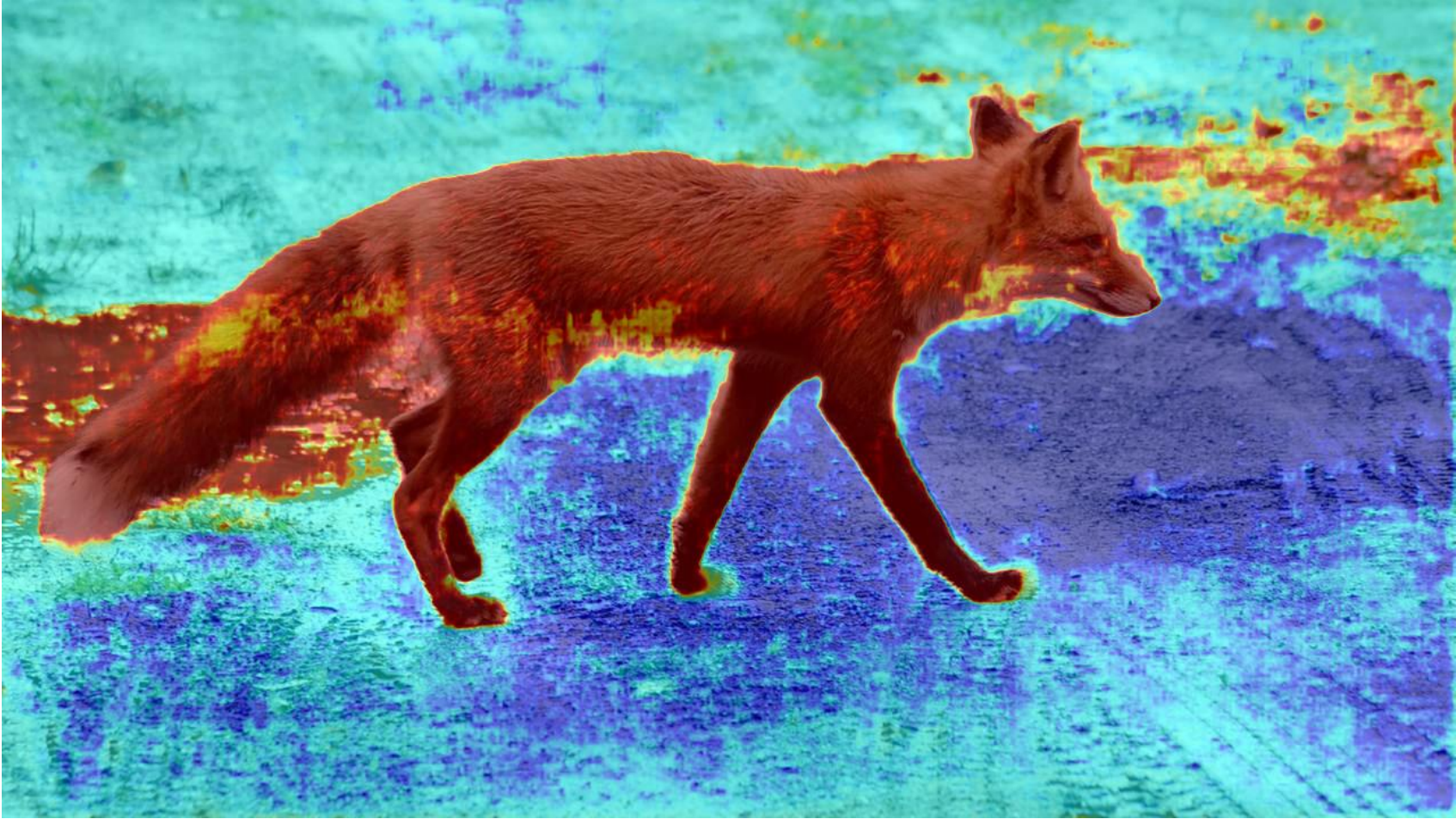} \\
        [-0.2cm]
        &
        \makebox[0.12\textwidth]{\scriptsize "fox"} &
        \makebox[0.12\textwidth]{\scriptsize "fox", "puddle"}  \\ [0.1cm]


        \includegraphics[width=0.23\textwidth]{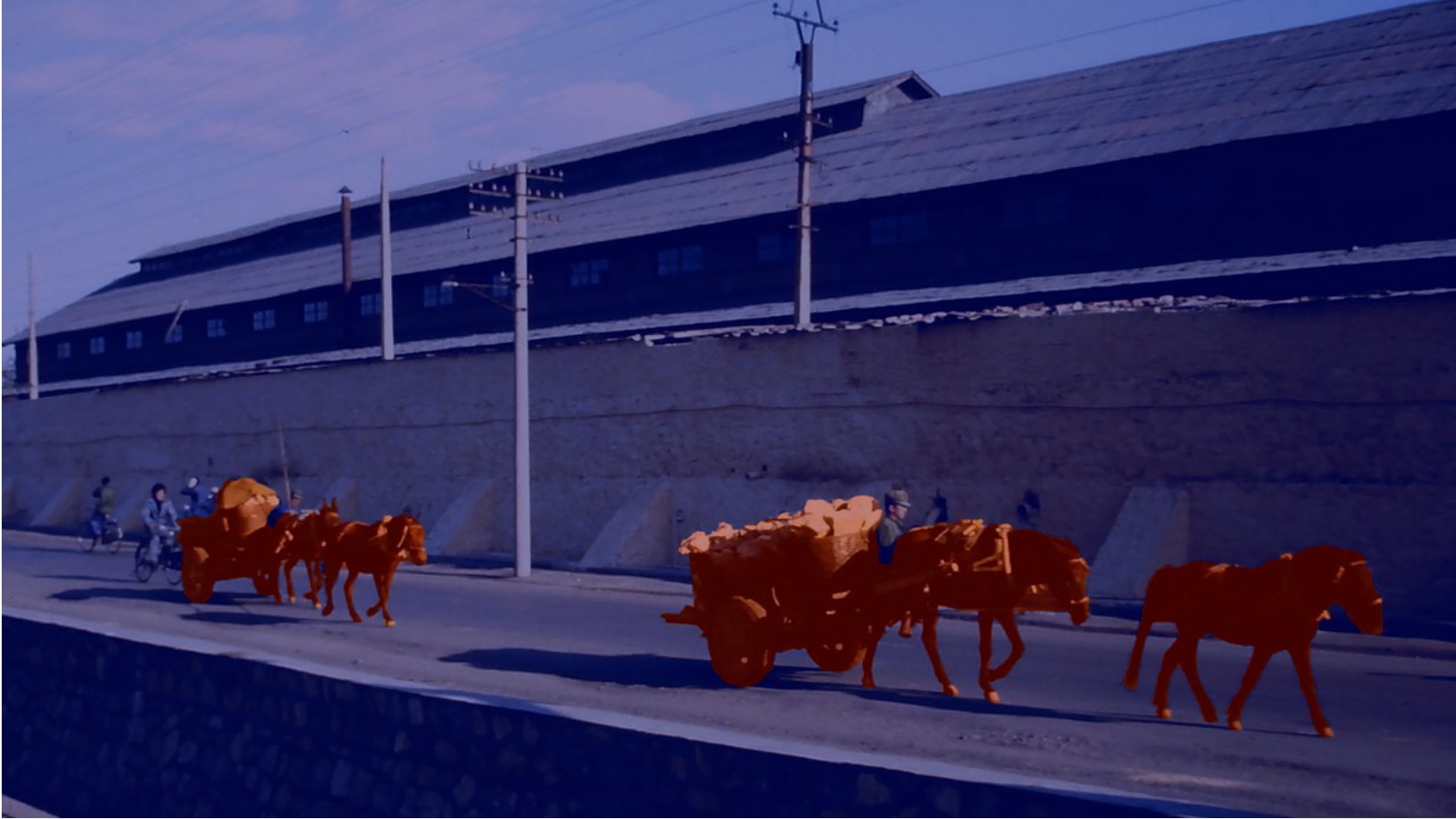} &
        \includegraphics[width=0.23\textwidth]{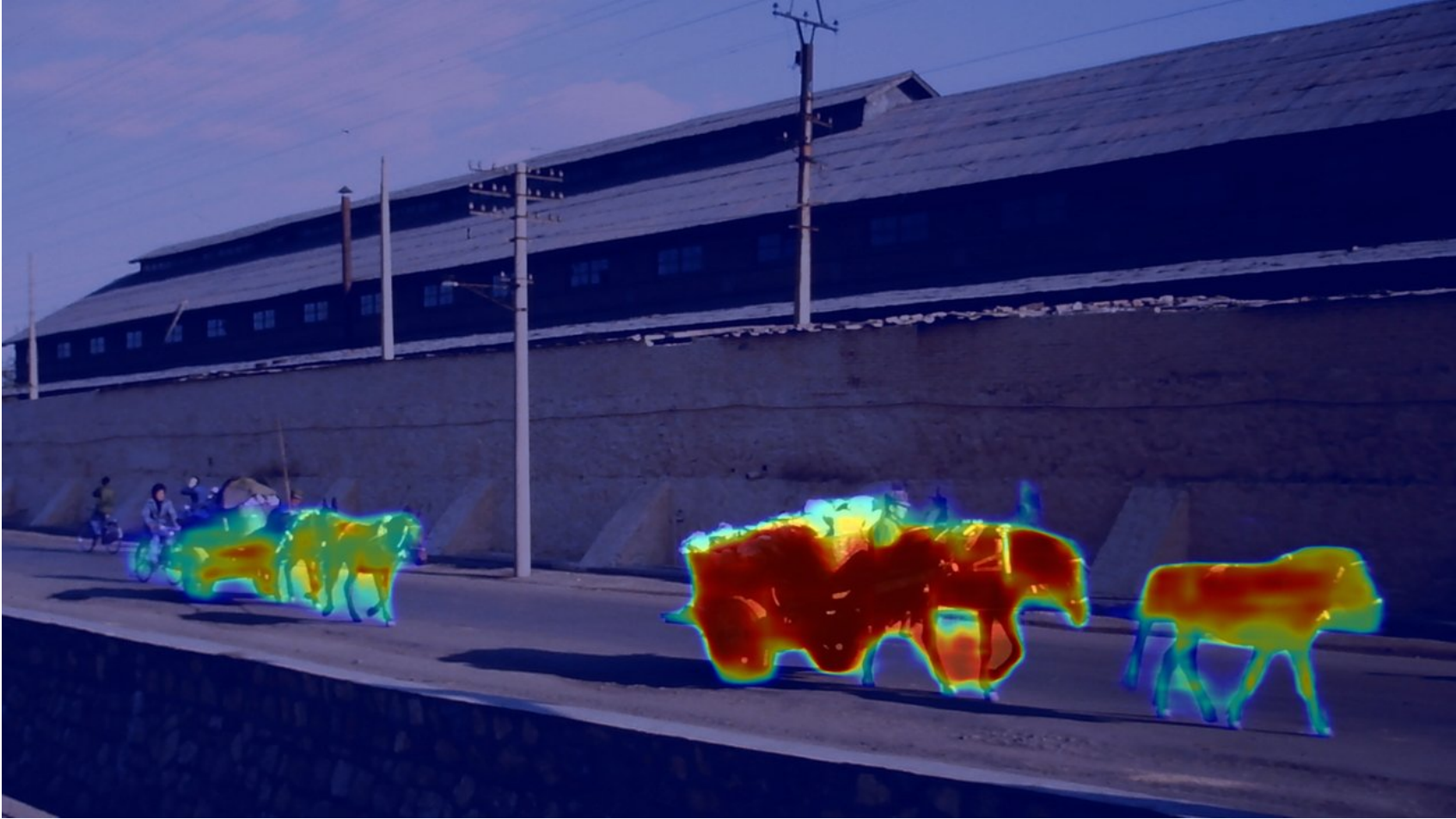} &
        \includegraphics[width=0.23\textwidth]{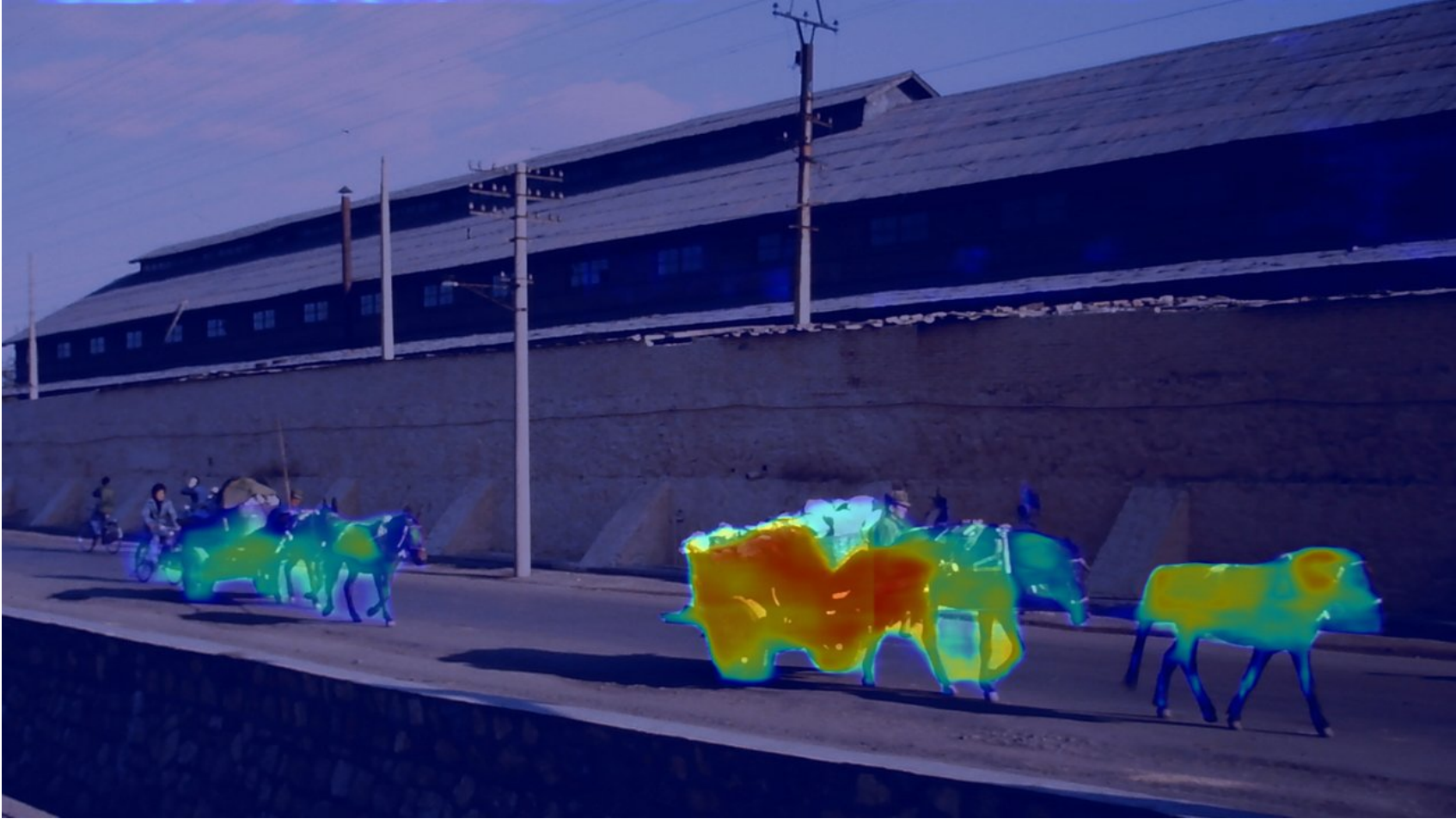} &
        \includegraphics[width=0.23\textwidth]{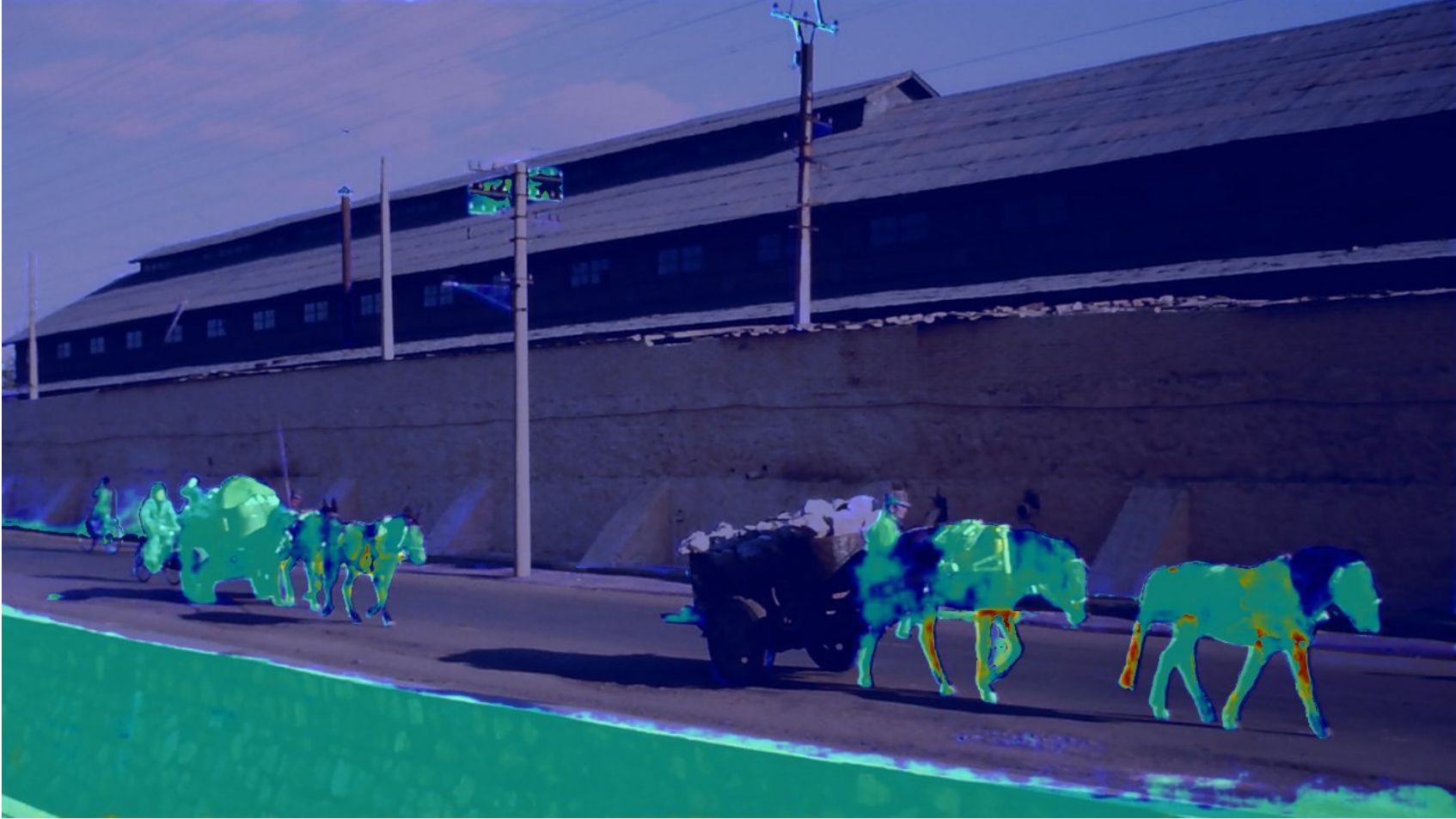} \\
        [-0.2cm]
        &
        \makebox[0.12\textwidth]{\scriptsize "horse cart"} &
        \makebox[0.12\textwidth]{\scriptsize "horse", "cart"}  \\
    \end{tabular}
    \caption{Qualitative anomaly segmentation results comparison of our method Clipomaly. For the columns that show our method, generated names for anomalies are shown below the image.}
    \label{fig:qualitative_comparison_anomaly}
\end{figure*}

Our RAM-based approach generates fewer false positive anomaly labels, benefiting from training on large-scale image–caption data. Figure \ref{fig:qualitative_comparison_anomaly} shows that CLIP-All and CLIP-Best yield segmentation-like heatmaps that are tighter and less noisy than Maskomaly, reflecting their origin from segmentation logits. CLIP-All is advantageous when anomalies span multiple semantic classes (e.g., donkey and hay), though it can also introduce false positives (e.g., fox and puddle). This behavior is linked to the quality of the initial unknown masks (Section \ref{sec:unknown}), and Table \ref{tab:textgen_match} confirms that CLIP-All outperforms CLIP-Best when ground-truth unknown masks are provided.

\subsection{Open-World Segmentation}
Since we have already shown that our method outperforms ContMAV in anomaly segmentation, the remaining question for Open World Segmentation is whether segmentation performance on known classes remains stable. Demonstrating that our pipeline maintains accuracy on known classes confirms its effectiveness for Open World Segmentation. Table \ref{tab:open_world} shows that our anomaly segmentation has only minimal impact on the mIoU of known classes. On Cityscapes~\cite{Cityscapes} performance remains stable, and on BDD-Anomaly~\cite{BDDAnomaly} we likewise preserve accuracy when training is restricted to the known classes of the dataset. 

\subsection{Generated Class Names}
\begin{figure*}[ht]
    \centering
    \resizebox{0.99\textwidth}{!}{%
    \begin{tabular}{@{}ccccc@{}}
        \includegraphics[height=2.5cm,width=0.25\textwidth]{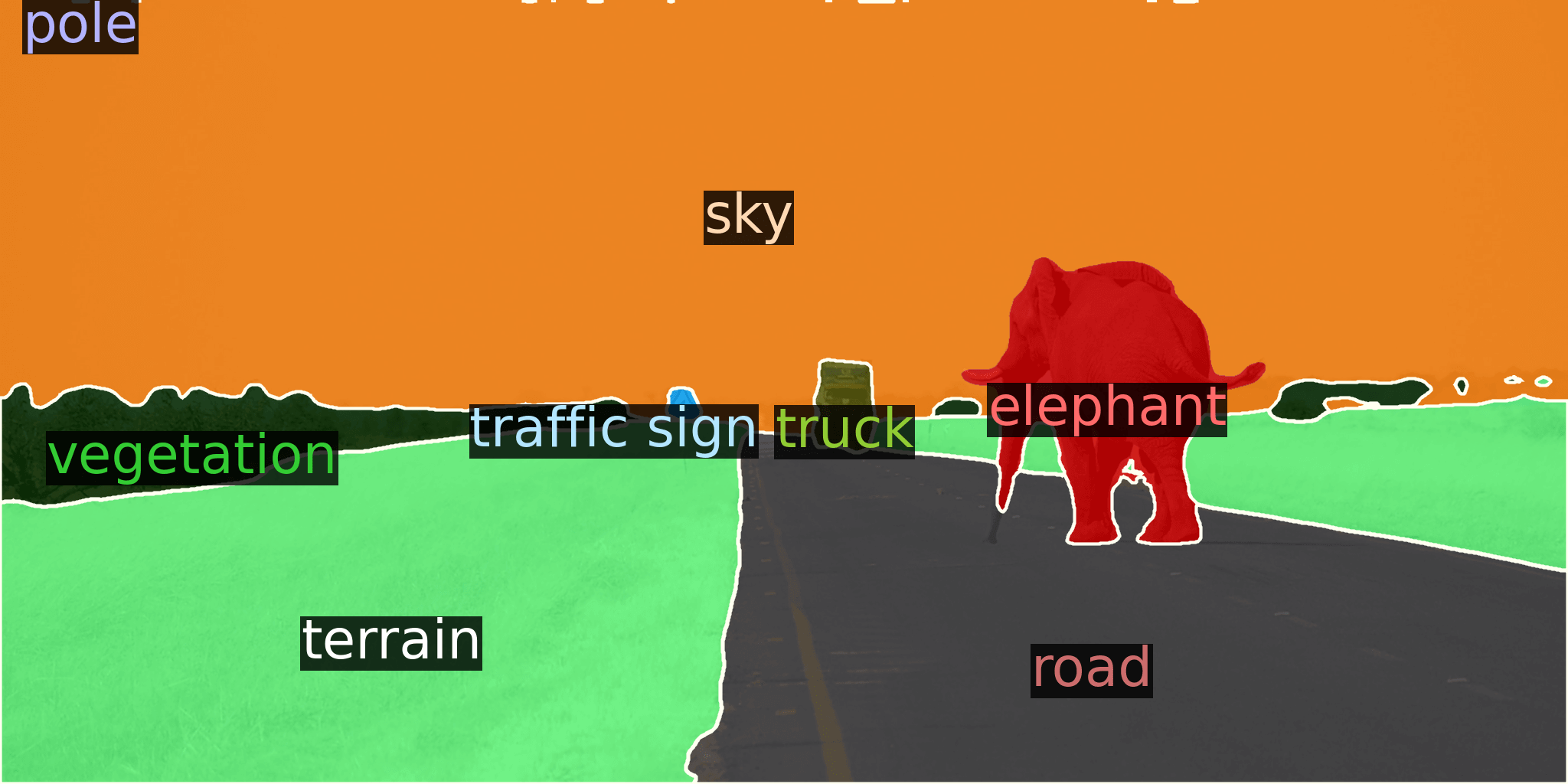} &
        \includegraphics[height=2.5cm,width=0.25\textwidth]{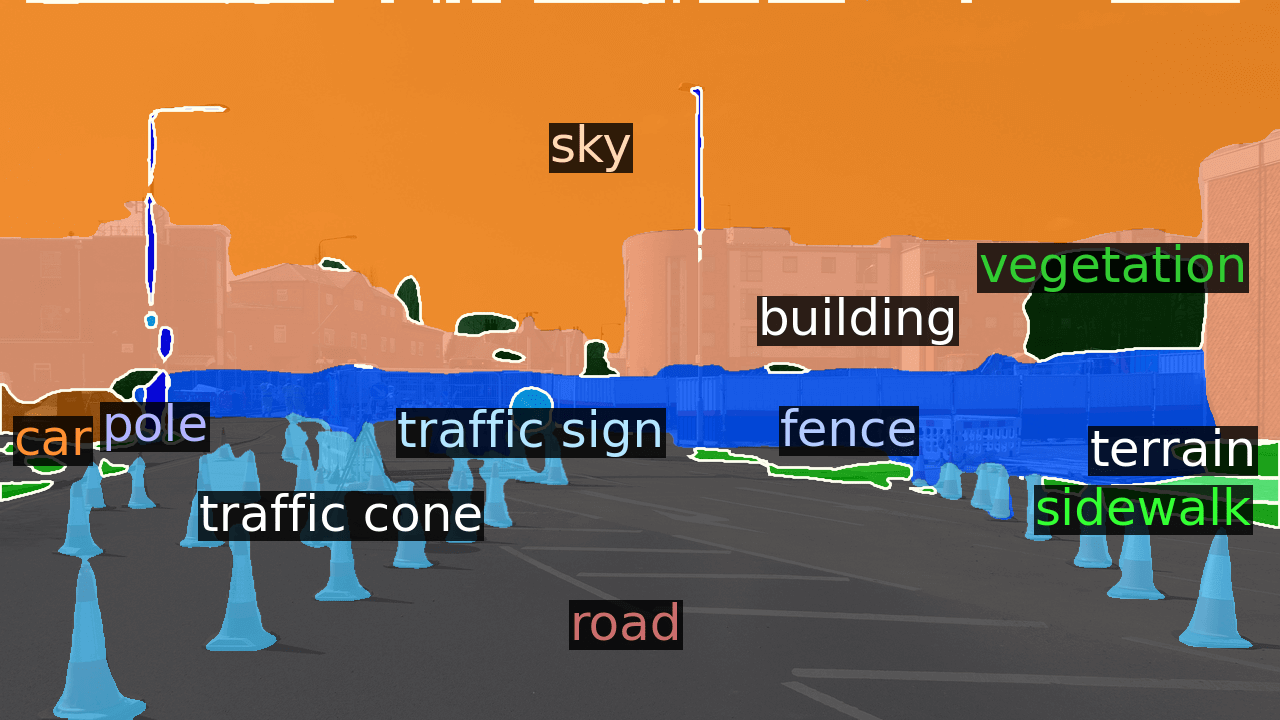} &
        \includegraphics[height=2.5cm,width=0.25\textwidth]{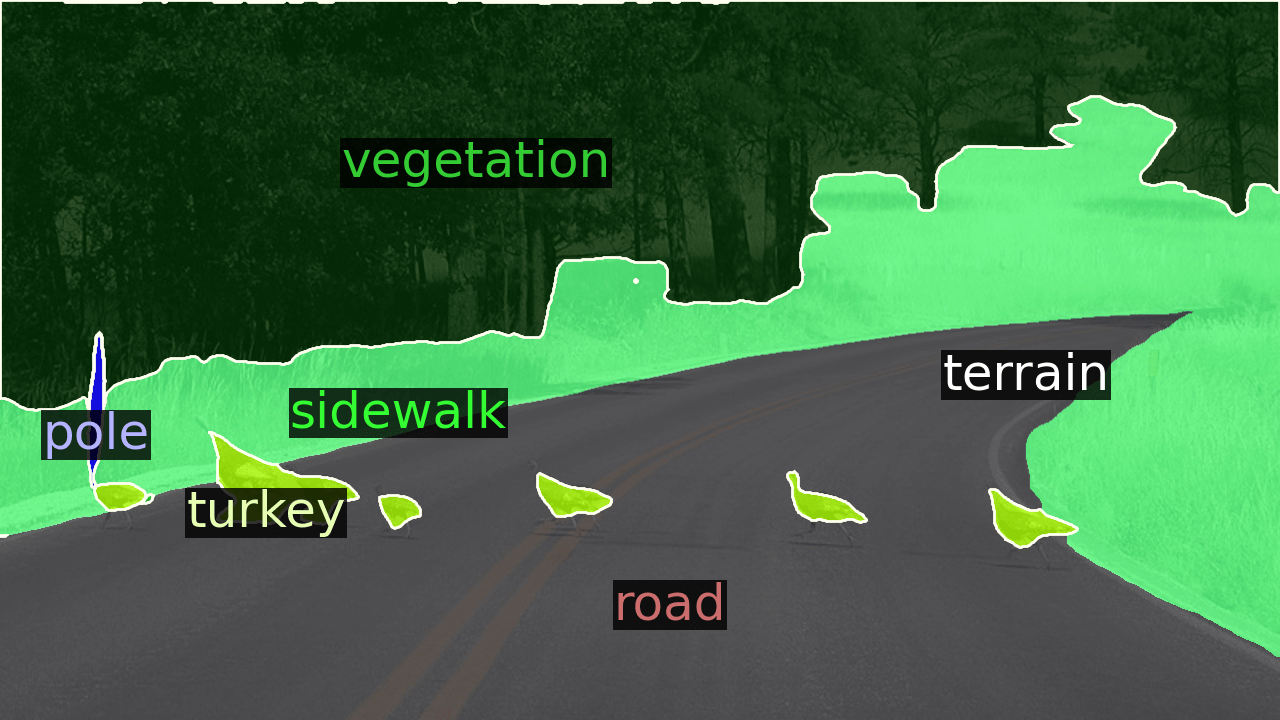} &
        \includegraphics[height=2.5cm,width=0.25\textwidth]{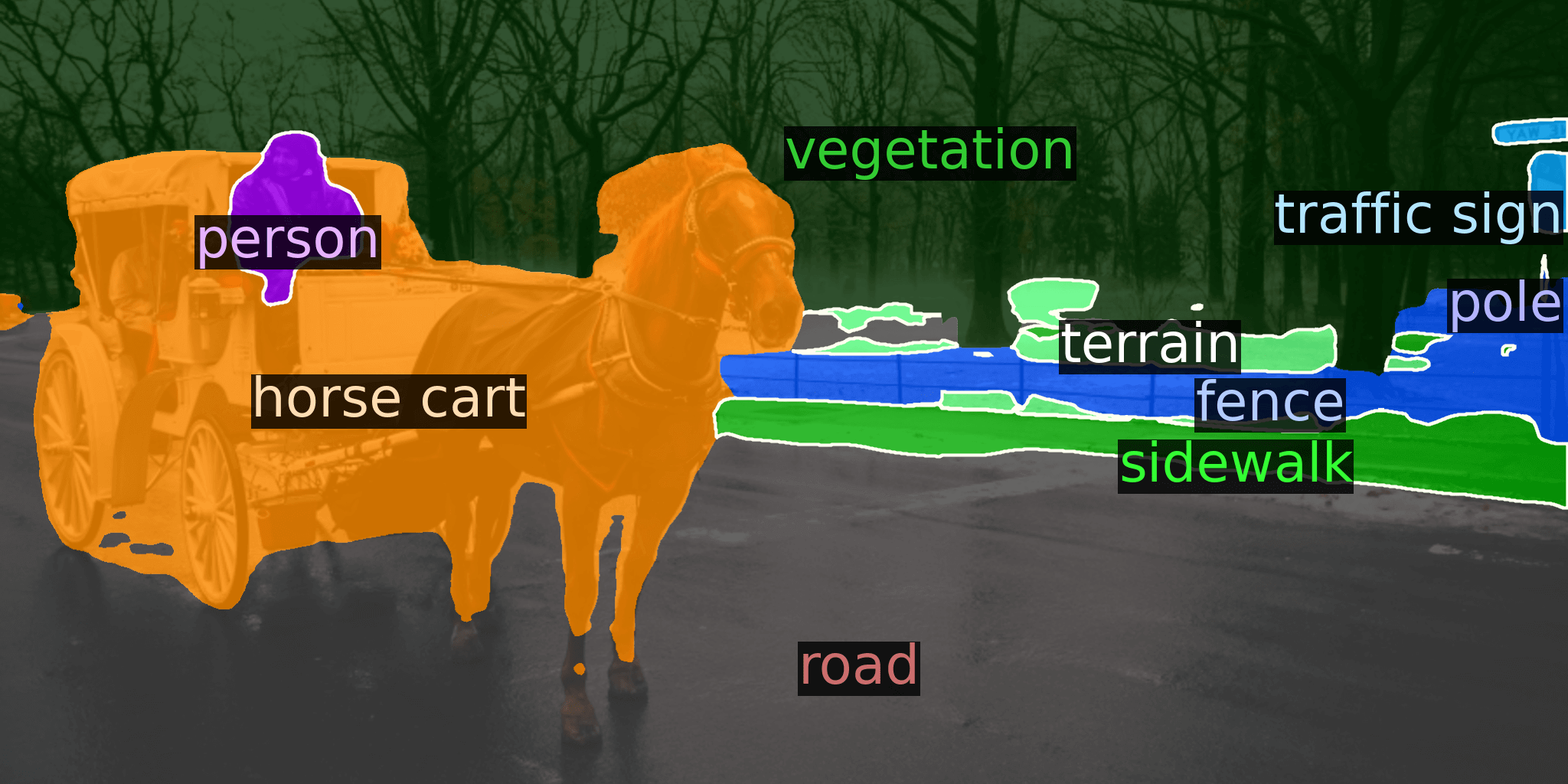} \\

        \includegraphics[height=2.5cm,width=0.25\textwidth]{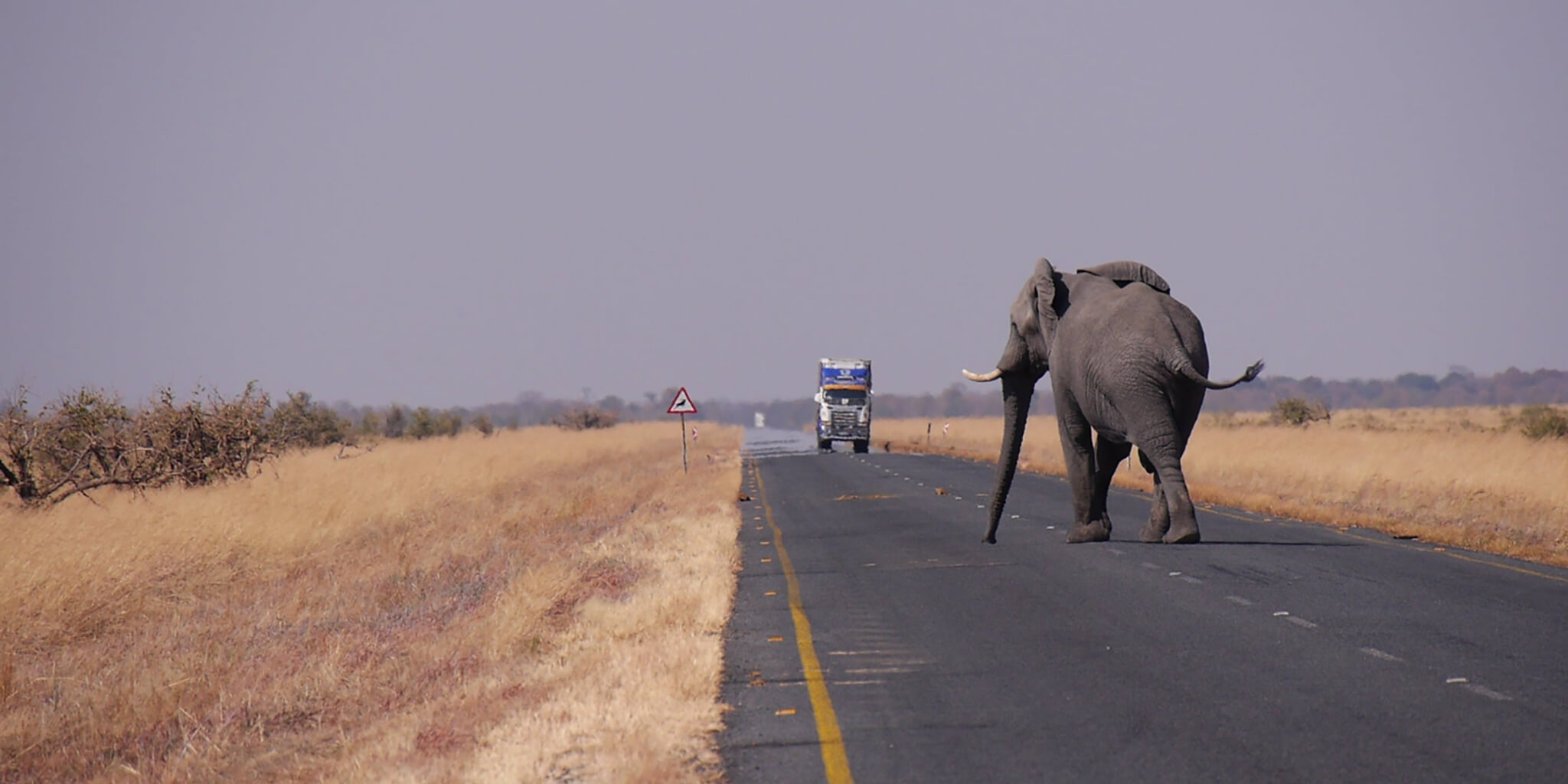} &
        \includegraphics[height=2.5cm,width=0.25\textwidth]{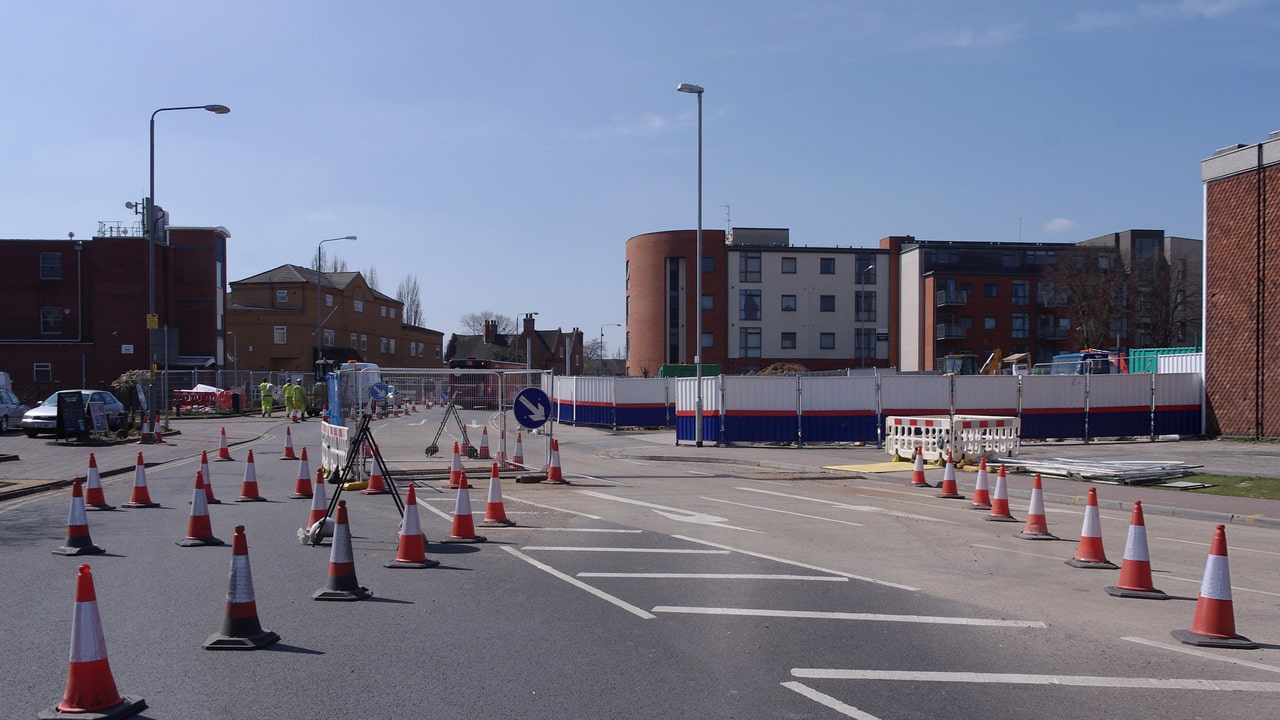} &
        \includegraphics[height=2.5cm,width=0.25\textwidth]{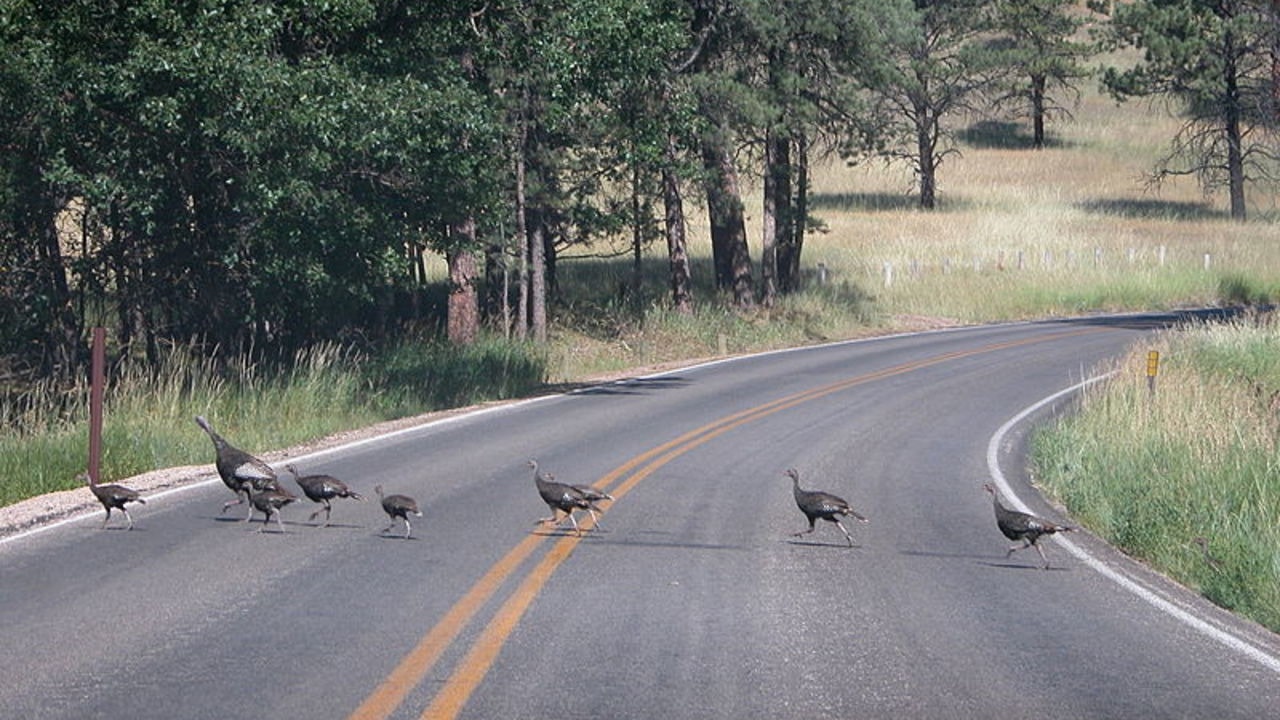} &
        \includegraphics[height=2.5cm,width=0.25\textwidth]{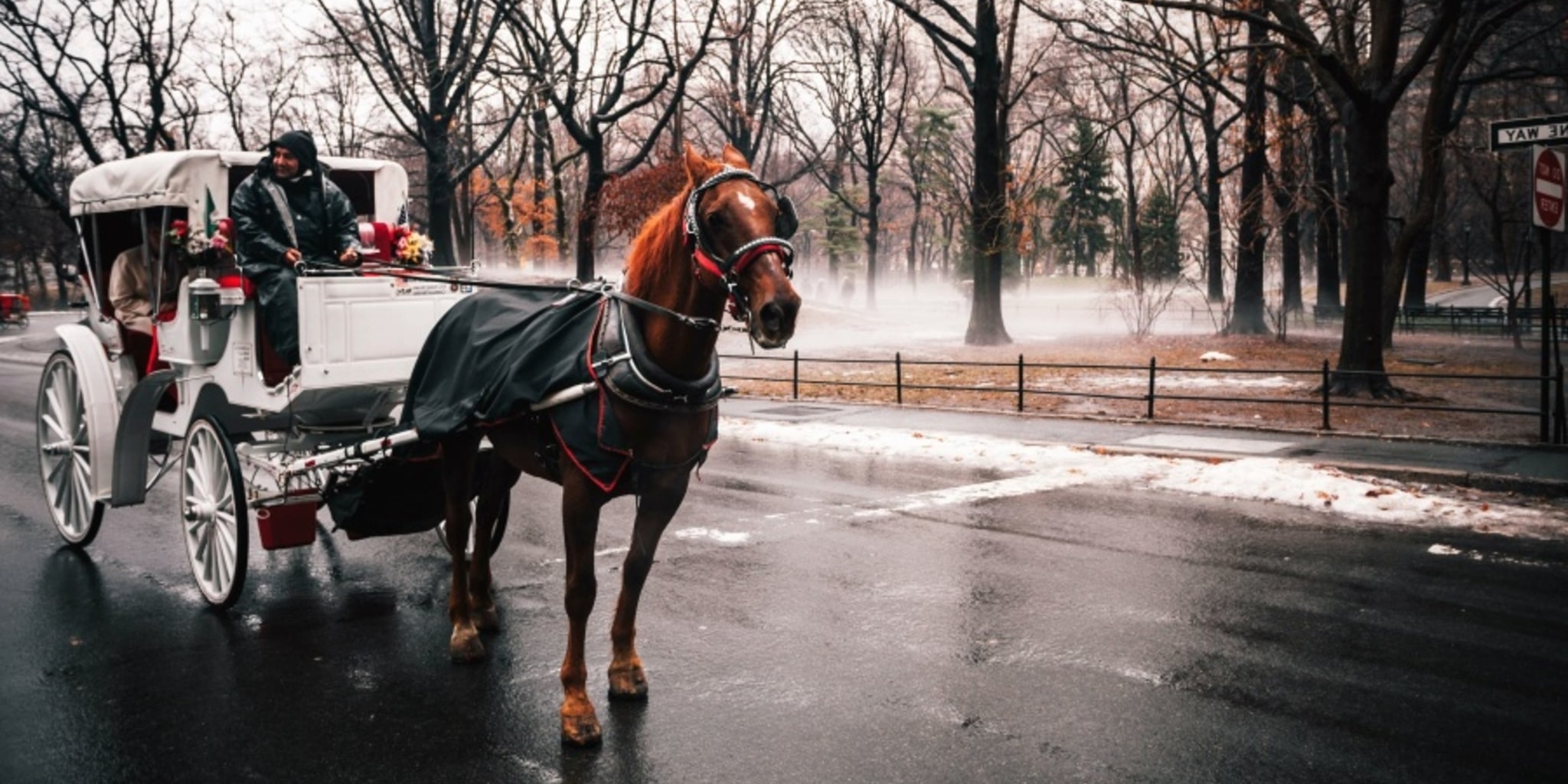} \\
    \end{tabular}
    }
    \caption{Qualitative segmentation results of \textbf{Ours-RAM} on the SMIYC AnomalyTrack dataset using our extended vocabulary method. The top row shows the predicted segmentations, while the bottom row shows the corresponding input images.}

    \label{fig:sota_qualitative_comparison}
\end{figure*}

We achieve reasonable performance in generating class names for anomalies. Table \ref{tab:names_quality} reports the Sentence-BERT similarity \cite{sBERT} between the generated names and the ground-truth labels that we manually annotated for RoadAnomaly. The labels are available at the following link: \url{https://***}. This metric has also been used in works on vocabulary-free semantic segmentation \cite{CASED}. Our RAM-based method shows strong semantic alignment with the annotated labels. Fig. \ref{fig:sota_qualitative_comparison} shows qualitative results which further underline this conclusion. Our dictionary-based methods exhibit lower Sentence-BERT similarity, partly because the generated names can be more detailed (e.g., ‘Indian elephant’ versus the more general annotation ‘elephant’), while the ground-truth labels tend to be broader.

\newlength{\firstImageHeight}

\begin{table}[h]
\centering
\caption{Comparison of open-world and closed-world setting on BDD-Anomaly~\cite{BDDAnomaly} and Cityscapes~\cite{Cityscapes} in mIoU.}
\label{tab:open_world}
\begin{tabular}{l l cc}
\toprule
\textbf{Method} & \textbf{Setting} & \multicolumn{2}{c}{\textbf{Dataset}} \\
\cmidrule(lr){3-4}
& & \textbf{BDD-Anomaly} & \textbf{Cityscapes} \\
\midrule
Ours     & CW & 64.2 & 69.8 \\
Ours     & OW & 62.4 & 69.6 \\
\bottomrule
\end{tabular}
\end{table}


\begin{table}[H]
\centering
\caption{Soft Jaccard Index with SBERT similarity on RoadAnomaly.}
\label{tab:names_quality}
\begin{tabular}{l l c}
\toprule
\textbf{Method} & \textbf{Tagging} & \textbf{RoadAnomaly} \\
\midrule
Ours & WordNet DB & 0.44  \\
Ours & RAM        & 0.69  \\
Ours & CaSED DB   & 0.5  \\
\bottomrule
\end{tabular}%
\end{table}

\section{Conclusion and Future Work}
We propose a novel open-world segmentation approach based on vision-language embeddings, which outperforms existing methods. A key advantage is dynamic vocabulary adaptation:  commonly detected anomalies can dynamically be added (temporarily or permanently) to the closed vocabulary, without continued training. This also prevents over-segmentation and conflicts between similar class names by including only the relevant classes for a specific deployment. 

A current limitation is the accuracy of the CLIP-based unknown similarity maps, which could be improved with an uncertainty loss. Potential future work includes extending CLIP-based mask-to-class matching to vocabulary-free semantic segmentation.


\section*{Acknowledgments}
This work was supported by BMW Group.

\begin{table*}[ht]
    \centering
    \caption{Summary table of our method trained on Cityscapes.}
    \label{tab:ours_better_cityscapes}
    \resizebox{0.97\textwidth}{!}{%
    \begin{tabular}{l cc ccccc ccccc}
        \toprule
        \multirow{2}{*}{\textbf{Method}} & \multirow{2}{*}{\textbf{Tagging}} & \multirow{2}{*}{\textbf{CLIP Match}} & \multicolumn{5}{c}{\textbf{RoadAnomaly}} & \multicolumn{5}{c}{\textbf{SMIYC}} \\
        & & & mIoU & AUPR & $\text{FPR90}$ & $\text{FPR95}$ & AUROC & mIoU & AUPR & $\text{FPR90}$ & $\text{FPR95}$ & AUROC \\
        \cmidrule(lr){4-8} \cmidrule(lr){9-13}
        Ours & RAM & Best & \textbf{57.8} & \textbf{84.74} & \textbf{6.84} & \underline{14.9} & \textbf{97.06} & \underline{75.1} & \underline{94.74} & 3.35 & 9.53 & 98.54 \\
        Ours & WordNet Dict & Best & \underline{53.0} & \underline{76.52} & 23.26 & 27.43 & 94.50 & \textbf{76.17} & 93.86 & \textbf{2.11} & \textbf{3.99} & \underline{98.89} \\
        Ours & CaSED Dict & Best & 49.42 & 71.29 & 27.90 & 35.47 & 93.08 & 75.6 & \textbf{95.48} & 3.35 & 9.53 & 98.81 \\
        Ours & RAM & All & 53.2 & 76.3 & 22.73 & 27.19 & 93.6 & 75.1 & 94.74 & 3.35 & 9.53 & -- \\ 
        \midrule
        Maskomaly~\cite{Maskomaly} & – & – & 45.04 & 70.9 & \underline{11.98} & \textbf{12.9} & \underline{95.38} & 67.5 & 93.4 & \underline{2.65} & \underline{6.9} & \textbf{98.93} \\
        \bottomrule
    \end{tabular}
    }
\end{table*}

\bibliographystyle{IEEEtran}
\bibliography{references}



\end{document}